\documentclass[lettersize,journal]{IEEEtran}
\usepackage{amsmath,amsfonts}
\usepackage{algorithm}
\usepackage{array}
\usepackage[caption=false,font=normalsize,labelfont=sf,textfont=sf]{subfig}
\usepackage{textcomp}
\usepackage{stfloats}
\usepackage{url}
\usepackage{verbatim}
\usepackage{cite}
\usepackage[colorlinks,linkcolor=red,citecolor=blue]{hyperref}

\ifCLASSINFOpdf
\usepackage[pdftex]{graphicx}
\DeclareGraphicsExtensions{.eps,.pdf,.tiff,.jpeg,.png}
\else
\usepackage[dvips]{graphicx}
\DeclareGraphicsExtensions{.eps}
\fi
\graphicspath{{pic/}{pic/authors}}

\usepackage{graphicx, algcompatible, epstopdf,epsfig,amssymb,multirow, enumerate,float, diagbox,makecell, colortbl}
\usepackage[caption=false]{subfig}

\newcommand{\mathcalX}{\mathcal{X}}
\newcommand{\mathcalY}{\mathcal{Y}}

\begin{document}
	
	\title{STGlow: A Flow-based Generative Framework with Dual Graphormer for Pedestrian Trajectory Prediction}
	\author{{Rongqin Liang,~\IEEEmembership{Student Member,~IEEE,} 
			Yuanman Li,~\IEEEmembership{Member,~IEEE,} 
			Jiantao Zhou,~\IEEEmembership{Senior Member,~IEEE,} and 
			Xia Li,~\IEEEmembership{Member,~IEEE}}
		\thanks{This work was supported in part by in part by the Key project of Shenzhen Science and Technology Plan under Grant 20220810180617001 and the Foundation for Science and Technology Innovation of Shenzhen under Grant RCBS20210609103708014;  in part by the Natural Science Foundation of China under Grant 62001304; in part by the Guangdong Basic and Applied Basic Research Foundation under Grant 2022A1515010645;  in part by the Open Research Project Programme of the State Key Laboratory of Internet of Things for Smart City (University of Macau) under Grant SKLIoTSC(UM)-2021-2023/ORP/GA04/2022.
			
			Rongqin Liang, Yuanman Li and Xia Li are with the Guangdong Key Laboratory of Intelligent Information Processing, College of Electronics and Information Engineering, Shenzhen University, Shenzhen, China. (Corresponding author: Yuanman Li, email: yuanmanli@szu.edu.cn.)
			
			Jiantao Zhou is with the State Key Laboratory of Internet of Things for Smart City, and also with the Department of Computer and Information Science, University of Macau. e-mail: jtzhou@um.edu.mo.}
	}
	\markboth{IEEE TRANSACTIONS ON NEURAL NETWORKS AND LEARNING SYSTEMS}%
	{Shell \MakeLowercase{\textit{et al.}}: A Sample Article Using IEEEtran.cls for IEEE Journals}
	\maketitle
	\begin{abstract}
		The pedestrian trajectory prediction task is an essential component of intelligent systems. Its applications include but are not limited to autonomous driving, robot navigation, and anomaly detection of monitoring systems. Due to the diversity of motion behaviors and the complex social interactions among pedestrians, accurately forecasting their future trajectory is challenging.
		Existing approaches commonly adopt GANs or CVAEs to generate diverse trajectories. However, GAN-based methods do not directly model data in a latent space, which may make them fail to have full support over the underlying data distribution; CVAE-based methods optimize a lower bound on the log-likelihood of observations, which may cause the learned distribution to deviate from the underlying distribution. The above limitations make existing approaches often generate highly biased or inaccurate trajectories.
		In this paper, we propose a novel generative flow based framework with dual graphormer for pedestrian trajectory prediction (STGlow). Different from previous approaches, our method can more precisely model the underlying data distribution by optimizing the exact log-likelihood of motion behaviors. Besides, our method has clear physical meanings for simulating the evolution of human motion behaviors. The forward process of the flow gradually degrades complex motion behavior into simple behavior, while its reverse process represents the evolution of simple behavior into complex motion behavior.
		Further, we introduce a dual graphormer combining with the graph structure to more adequately model the temporal dependencies and the mutual spatial interactions.
		Experimental results on several benchmarks demonstrate that our method achieves much better performance compared to previous state-of-the-art approaches.
		
	\end{abstract}
	
	\begin{IEEEkeywords}
		Generative flow, trajectory prediction, graph learning, attention mechanism, deep neural network.
	\end{IEEEkeywords}
	

	\section{Introduction}
	Trajectory prediction, as one of the most important future behavior modeling tasks, aims to predict the future trajectory based on the observed trajectory. It plays an important role in applications such as self-driving vehicles \cite{liang2019peeking}, autonomous navigation robots \cite{SzemesTIM2005}, anomaly behavior detection \cite{musleh2010identifying, 9749781}, video surveillance \cite{BastaniTIP2016,LinTIP2012,JiangTIP2009} and so on.
	Despite that significant advances have been achieved recently \cite{gupta2018social,mohamed2020social,yu2020spatio,liang2021temporal,yuan2021agentformer,Li2022,gu2022stochastic, 9667184}, accurately predicting future trajectories of pedestrians remains challenging due to the inherent properties of pedestrians. 
	First, due to the differences of human intent and unique behavior patterns, the future trajectories of pedestrians are full of diversity, even when they share the same historical trajectory. Second, influenced by surrounding agents, there are highly complex social interactions among pedestrians, which drive pedestrians to make decisions such as walking parallel, walking in groups, or changing direction / speed to avoid collisions. Faced with the challenge of diverse future trajectories, most previous works \cite{gupta2018social,salzmann2020trajectron++, liang2021temporal,yuan2021agentformer,gu2022stochastic,Li2022} applied generative models to model the multi-modality of human motion behaviors. For instance, some studies \cite{gupta2018social,Li2022,liang2021temporal} employed generative adversarial networks (GANs) \cite{goodfellow2014generative} to predict the distribution of future trajectories. 
	However, GAN-based methods do not directly model data in a latent space, which may make them fail to have full support over the underlying data distribution, thus generating highly biased trajectories. In addition, the training process of GANs is often unstable due to the adversarial learning.
	Alternatively, some works exploited conditional variational auto-encoders (CVAEs) \cite{mangalam2020not,salzmann2020trajectron++,mangalam2021goals,yao2021bitrap,chen2021personalized} or diffusion model \cite{gu2022stochastic} to model the diversity of future trajectories. However, 
	both of them optimize the variational lower bound on the log-likelihood of observations \cite{dinh2014nice}, which may cause the learned distribution to deviate from the underlying distribution. This means that using a lower bound criterion may yield a suboptimal solution with respect to the true log-likelihood, resulting in inaccurate future trajectories. Therefore, how to more precisely model the underlying distribution of pedestrian trajectories is very important for pedestrian trajectory prediction tasks.
	
	To straightforwardly model the social interactions among pedestrians, some researches \cite{mohamed2020social,shi2021sgcn, 9447207, 9728758} proposed to represent the social interactions between pedestrians utilizing the topology of graphs. However, graph-based methods may suffer from over-smoothing problems on node features \cite{ying2021transformers}, which means that when constructing graphs for crowded environments, the features of nodes will be smoothed by the aggregation of nodes. This may lead to the loss of unique behavior characteristics of pedestrians.
	Therefore, how to effectively model the social interactions while maintaining unique behavioral features of pedestrians is still a challenge for pedestrian trajectory prediction tasks.
	
	Faced with the above challenges on diverse trajectories and social interactions, in this paper, we propose a novel flow-based generative framework with dual graphormer for pedestrian trajectory prediction (STGlow).
	Firstly, to more precisely model the underlying distribution, we propose a generative flow framework with pattern normalization (Glow-PN) to produce multiple reasonable future trajectories. In contrast to previous GAN-based and CVAE-based models, our framework optimizes the exact log-likelihood of observations by mapping a complex data distribution into a simple and tractable one through a series of invertible transformations. Similar to the process of artists creating artworks, complex motion behaviors of pedestrians are not accomplished at one stroke, but based on simple motion behaviors (\textit{e.g.}, every step of walking or every action) combined with individual behavior habits originating from various `biases' such as people's walking habits, traffic awareness, and potential intentions. Hence, we propose to characterize the evolution of motion behavior from simple to complex by modeling the `biases' progressively utilizing the generative flow with a series of simple and tractable functions.
	Secondly, to model the social interactions more effectively, we further propose a dual graphormer to extract the representations of motion behaviors and model the temporal dependencies and the mutual spatial interactions. The proposed dual graphormer combines graph structure with transformers, where the designed attention mechanism can adaptively focus on all other nodes. Our method not only enables intuitive and effective modeling of social interactions, but also greatly alleviates the problem of over-smoothing of nodes. Specifically, in the training phase, a Glow-PN is applied to learn the distribution of the motion behavior conditioned on the representation of social interactions. In the inference phase, simple behaviors are sampled from the standard normal distribution and formed into ``evolved” representations of complex motion behaviors using the reverse process of Glow-PN conditioned on the representation of social interactions. These representations of complex motion behaviors are eventually decoded into predicted trajectories through a bidirectional trajectory prediction module.
	The main contributions of our work can be summarized as follows:
	\begin{enumerate}
		\item We present a novel diverse trajectory prediction framework based on generative flow, simulating the evolution of human motion behaviors from simple to complex. In contrast to previous approaches, our method can more precisely model the underlying distribution by optimizing the exact log-likelihood of motion behaviors.
		Besides, a pattern normalization is carefully developed to normalize the unique behavior pattern of pedestrians, greatly improving the prediction accuracy.
		\item We further propose a dual graphormer to extract the representations of motion behaviors and model the social interactions in both temporal and spatial domains. Different from previous Transformer based methods, our dual graphormer combined with the graph structure more adequately models the temporal dependencies and the mutual spatial interactions.
		\item The proposed framework achieves state-of-the-art performance on widely used pedestrians trajectory prediction benchmarks, providing a promising direction for generating diverse and reasonable trajectories. 
	\end{enumerate}
	
	The remainder of this paper is organized as follows. Section \ref{sec:related work} gives a brief review of related works. Section \ref{sec:proposed} details our proposed STGlow for pedestrian trajectory prediction. Extensive experimental results are presented in Section \ref{sec:experiment}, and we finally draw a conclusion in Section \ref{sec:conclusion}.

	\section{Related Works}\label{sec:related work}
	\subsection{Pedestrian Trajectory Prediction}
	Traditional trajectory prediction methods mainly rely on designing handcrafted rules to model human interactions
	\cite{helbing1995social,wang2007gaussian,tay2008modelling,mehran2009abnormal, pellegrini2009you}. For instance, Social Force \cite{helbing1995social} introduced attractive and repulsive forces to avoid collisions. Although these methods demonstrate the importance of interaction modeling, they are limited by the handcrafted features and perform poorly in trajectory prediction.
	
	With the great success of deep neural networks, the Recurrent Neural Network (RNN) and its variants are widely applied in trajectory prediction task \cite{alahi2016social,9031707,zhang2019sr,liang2019peeking} and motion prediction task \cite{Martinez_2017_CVPR,9321130}, on the basis of their good performance on sequence learning \cite{chung2015recurrent,graves2014towards,vinyals2015show}. Wherein, Social-LSTM \cite{alahi2016social} employed a Long Short-Term Memory (LSTM) to encode pedestrian trajectory and designed a Social Pooling to aggregate the global representation of neighboring pedestrians. 
	To enhance the representation ability of social interaction features, many studies \cite{hu2020collaborative,liang2019peeking,xu2020cf,zhang2019sr} have followed this idea of transmitting information between pedestrians and proposed different effective message passing mechanisms.
	Although the RNN-based approaches approach the trajectory prediction task in a data-driven manner, they ignore the important fact that the future trajectories of pedestrians are full of diversity due to the differences of human intents and unique behavior patterns.
	
	Besides, graph networks are utilized in various tasks such as action understanding \cite{9782720,9954217}, recommendation systems \cite{3219890}, and text classification \cite{Yao_Mao_Luo_2019}, due to their capability of modeling non-euclidean structured data. Recently, the intuitive modeling power of graph models has been applied to represent complex social interactions among pedestrians \cite{vemula2018social,9447207,huang2019stgat,mohamed2020social,bae2021disentangled,shi2021sgcn}.
	For instance, 
	the work GTPPO \cite{9447207} explored a social graph attention module that combines specific obstacle avoidance experiences (OAEs) with the graph attention to capture pedestrians' social interactions.
	DMRGCN \cite{bae2021disentangled} proposed a disentangled multi-scale aggregation to better represent social interactions between pedestrians on a weighted graph. Though the topology of graphs seems to be a straightforward way to represent social interactions, graph-based methods may suffer from over-smoothing problems on node features \cite{bae2021disentangled, ying2021transformers}, which may lead to the loss of the unique behavior characteristics of pedestrians.

	To generate diverse future trajectories, some researchers suggested employing generative models to model the diversity of human motion behaviors \cite{gupta2018social,zhao2019multi,Li2019IROS}.
	Part of these works were based on GANs \cite{gupta2018social,Li2022,liang2021temporal}. Among them, Social-GAN \cite{gupta2018social} applied GANs for the first time to generate diverse future trajectories and designed a pooling module to aggregate social interactions. Furthermore, TPNMS \cite{liang2021temporal} proposed a temporal pyramid structure to model both global and local contexts of human motion behaviors. Another part of the works \cite{mangalam2020not,salzmann2020trajectron++,yuan2021agentformer,chen2021personalized,yao2021bitrap} applied CVAEs to explicitly encode the distribution of diverse future trajectories. For instance, Trajectron++ \cite{salzmann2020trajectron++} utilized the latent variable framework of CVAEs to explicitly encode diversity and modeled social interactions in combination with a graph-structured recurrent model, while PECNet \cite{mangalam2020not} embedded the distant trajectory endpoints into a latent space to assist in long-range diverse trajectory prediction. More recently, MID \cite{gu2022stochastic} devised a Transformer-based diffusion model for trajectory prediction with a reverse process of motion indeterminacy diffusion. Though previous generative models have achieved promising performance in modeling the diversity of human behaviors, these approaches still have inherent limitations, \textit{e.g.},
	methods based on GANs may not fully support the data distribution due to the lack of encoding latent variables, while methods based on CVAEs and diffusion models optimize a lower bound on the log-likelihood of observations. Such limitations could make them generate biased or inaccurate trajectories.
	
	\subsection{Normalizing Flow}
	Normalizing Flows (NFs) are invertible generative models that map complex data distributions to simple and tractable ones.
	Recently, NFs have been successfully applied to a variety of generation tasks such as image generation \cite{dinh2014nice,dinh2016density,kingma2018glow,lugmayr2020srflow}, video generation \cite{kumar2019videoflow}, speech synthesis \cite{prenger2019waveglow,kim2018flowavenet}. For example, Glow \cite{kingma2018glow} proposed an invertible $ 1 \times 1 $ convolution for the generative flow and showed the efficiency of realistic-looking synthesis and manipulation of large images. WaveGlow \cite{prenger2019waveglow} proposed a flow-based network for generating high-quality speech from mel-spectrograms. 
	In our task, we propose a flow-based method for pedestrian trajectory prediction. Our proposed scheme has clear physical meanings to simulate the evolution of human motion behaviors.
	\begin{table*}[!t]
		\renewcommand{\arraystretch}{1.0}
		\centering  
		\fontsize{9}{9}\selectfont  
		\caption{A summary of the denotations.}
		\label{tb:denotation}
		\begin{tabular}{c|c|c|c}
			\hline
			Denotations & Descriptions & Denotations & Descriptions \\ \hline
			$\mathcalX$ & observed trajectories &  $R_{i}^{t} / SE$ & the embedding of the relative position \\
			$\mathcalY$ & future trajectories & $S_{i}^{t} / HE$ & the embedding of the relative steering angle \\
			$TH_i^{1:T}$ & the temporal embedding & $SG$ & the spatial graphormer \\
			$G_{tmp}$ & the temporal graph & $MB_{traget}$ & the representation of motion behavior \\
			$V_{tmp}$ & nodes of the $G_{tmp}$ & $ST_{traget}$ & the representation of social interaction\\
			$A_{tmp}$ & the adjacency matrix of the $G_{tmp}$ & $PN$ & pattern normalization \\
			$C_{i}^{t} / CE$ & the centrality embedding & $BiD$ & the bidirectional decoder \\
			$P_{i}^{t} / PE$ & the positional embedding & $\hat{G}^{t_p}$ & the goal of each pedestrian \\
			$TG$ & the temporal graphormer & $\hat{Y}_{F}^{t}$ & the predicted forward trajectory \\
			$SH_{1:N}^{t}$ & the spatial-temporal embedding & $\hat{Y}_{B}^{t_b}$ & the predicted backward trajectory \\
			$G_{spa}$ & the spatial graph  & $\hat{Y}_{Both}^{t_b}$ & the predicted bidirectional trajectory\\
			$V_{spa}$ & nodes of the $G_{spa}$ & $L_p$ & the loss of Glow-PN\\
			$A_{spa}$ & the adjacency matrix of the $G_{spa}$ & $L_{traj}$ & the loss of the bi-directional trajectory prediction \\
			\hline
		\end{tabular}
	\end{table*}
	\subsection{Transformer}
	Transformer \cite{vaswani2017NIPS}, which relies entirely on self-attention mechanisms to model global dependencies of the serialization inputs, has recently made remarkable progress in a variety of natural language processing (NLP) tasks \cite{brown2020language}, vision tasks \cite{dosovitskiy2020image,yuan2021tokens,you2021transformer, 9920222} and speech recognition \cite{9755926}. For instance, in vision tasks \cite{9359362}, ViT \cite{dosovitskiy2020image} sequentialized the image into a series of tokens and modeled the global dependencies of the image through the Transformer encoder. More recently, some works \cite{yu2020spatio,yuan2021agentformer,gu2022stochastic} have applied Transformer to pedestrian trajectory prediction tasks. Among them, STAR \cite{yu2020spatio} employed a temporal transformer and spatial transformer respectively to extract temporal dependencies and spatial interactions, while AgentFormer \cite{yuan2021agentformer} exploited an agent-aware Transformer to learn representations from both temporal and spatial dimensions. Different from prior works, in this work, we devise a dual graphormer, which can  more adequately model the temporal dependencies and the mutual spatial interactions.
	
	\begin{figure*}[!t]
		\centering
		\includegraphics[width=0.90\linewidth]{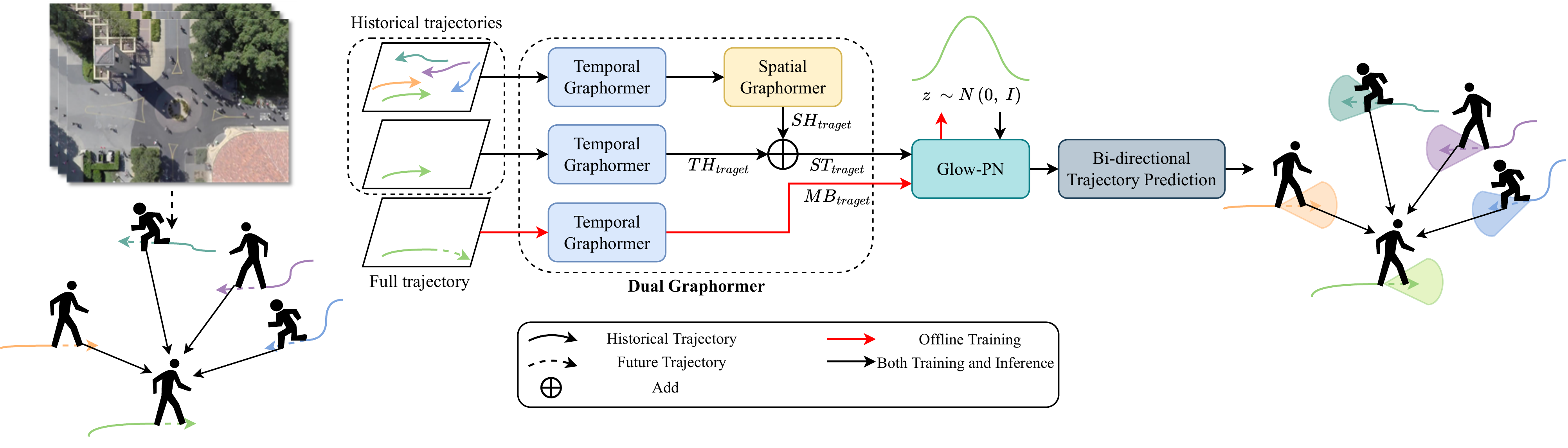}
		\caption{The framework of our STGlow algorithm. STGlow primarily consists of a dual graphormer, a generative flow with pattern normalization (Glow-PN) and a bi-directional trajectory prediction module. 1) First, the full trajectory of the target pedestrian is encoded into the representation of complex motion behavior through the temporal graphormer, while the historical trajectories are encoded into the representation of social interaction through the temporal and spatial graphormer; 2) then, a Glow-PN is applied to learn the distribution of complex motion behaviors using a series of simple reversible transformation, conditioned on the social interactions; 3) simple behaviors sampled from the standard normal distribution evolve into complex motion behaviors through the reverse process of Glow-PN, which are eventually fed into the bi-directional trajectory prediction module to predict diverse future trajectories.}
		\label{fig:framework}
	\end{figure*}
	\section{Proposed Approach: STGlow} \label{sec:proposed}
	The overall framework of STGlow model is illustrated in Fig. \ref{fig:framework}. It primarily consists of three components: 1) a dual graphormer to extract the representations of motion behaviors and model the temporal dependencies and the mutual spatial interactions; 2) a generative flow with pattern normalization (Glow-PN) to learn the underlying distribution of complex motion behaviors, conditioned on the social interactions; 3) a bi-directional trajectory prediction module to forecast diverse future trajectories. To facilitate check, we assemble the primary denotations and their accompanying explanations in Table \ref{tb:denotation}.
	
	\begin{figure*}[!t]
		\centering
		\includegraphics[width = 0.67\textwidth]{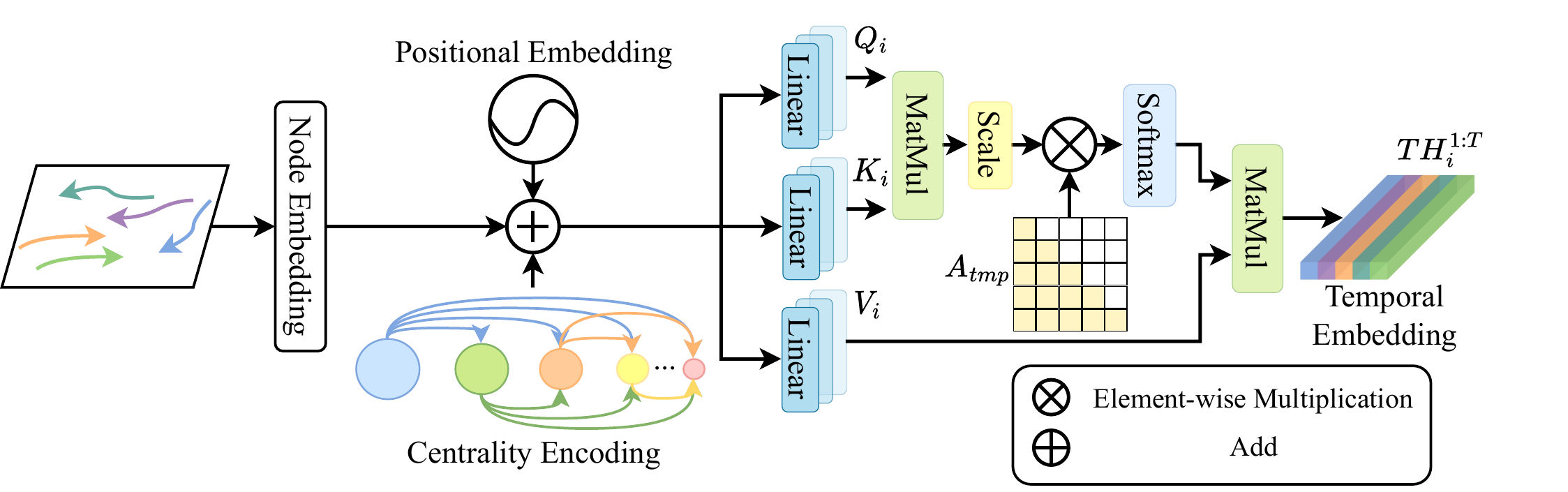}
		\caption{Illustration of the temporal graphormer.}	\label{fig:TG}
	\end{figure*} 
	\subsection{Problem Formulation}
	Given $N$ pedestrians with observed trajectories $\mathcalX = \{X_1^{(-t_o+1:0)},...,X_N^{(-t_o+1:0)}\}$ from time steps $T_{-t_o+1}$ to $T_0$ in the scene, the trajectory prediction algorithm aims to predict the future trajectories $\mathcalY = \{Y_1^{(1:t_p)},...,Y_N^{(1:t_p)}\}$ of all pedestrians in the upcoming time steps $T_1$ to $T_{t_p}$, where $X_i^t = (x_i^t, y_i^t) \in \mathbb{R}^{2}$ is the position of the $i$-th pedestrian at the time step $t$. The trajectory prediction algorithm takes as input the observed trajectories with $t_o$ time steps of all pedestrians in a scene, and aims to predict their future trajectories in the next $t_p$ time steps by a model $f\left( \cdot \right)$, denoted by
	\begin{equation}
	\hat{\mathcalY} = f\left(\mathcalX;W^*\right),
	\end{equation}
	where $\hat{\mathcalY}$ is the set of future trajectories predicted by $f(\cdot)$ and $W^*$ represents the set of learnable parameters in the model.

	For the sake of brevity, we hereafter drop the superscript when there is no ambiguity, \textit{i.e.}, $X_i \triangleq X_i^{(-t_o+1:0)}$ and $Y_i \triangleq Y_i^{(1:t_p)}$. We further use $X, Y$ to represent a generic history trajectory and the corresponding future trajectory, respectively.

	\subsection{Dual Graphormer}\label{sec:STG}
	Influenced by surrounding agents, highly complex social interactions among pedestrians may force them to make decisions such as walking parallel, walking in groups, or changing direction and speed to avoid collisions. Obviously, such social interactions contain both temporal dependencies and spatial interactions, which are fundamentally important for accurately predicting trajectories. In this work, we design a dual graphormer to more adequately model the temporal dependencies and the mutual spatial interactions. As shown in Fig. \ref{fig:framework}, our dual graphormer primarily consists of two components: 1) a temporal graphormer, and 2) a spatial graphormer.
	

	\subsubsection{Temporal Graphormer} \label{sec:TG}
	In this work, we decouple temporal dependencies into the  behavior-independent temporal dependencies and behavior-dependent temporal dependencies. The former type of dependencies reveals the importance of each previous time step to the future trajectory. The other type of dependencies models the relationships among different previous motion behaviors across the temporal domain. 
	

	Assume that there are $T$ time steps of each trajectory, denoted by $X_i^{1:T}$. As shown in Fig. \ref{fig:TG}, the temporal graphormer takes $X_i^{1:T}$ as input and outputs a set of embeddings $TH_i^{1:T}$ with temporal dependencies, which we define as \textbf{\textit{Temporal Embedding}}. Specifically, we first construct a temporal graph by treating each time step as a node of the graph,
	\begin{align}
	G_{tmp} = (V_{tmp}, A_{tmp}).
	\end{align} 
	Here $V_{tmp}=\{V_t|t=1,...,T\}$ represents nodes of $G_{tmp}$.  $A_{tmp}$ is the adjacency matrix describing the temporal dependencies, i.e., the motion behavior at a given time can only be affected by previous motion states rather than future motion states. Based on this fact, we define $A_{tmp}$ as
	\begin{equation}
	A_{tmp}^{ij}=\left\{ \begin{array}{l}
	1,\,\,\,\,\,\,\,\,\,\,\,\,\,\,\,\,\,\,\,\,\,\,\,\,\,\,\,\,\,\,\,i\ge j\\
	-inf,\,\,\,\,\,\,\,\,\,\,\,\,\,\,\,\,\,\,\,\,i<j\\
	\end{array},\ i,j\in \left[ 1,T \right] \right..
	\end{equation} 
	
	To model the behavior-independent temporal dependencies, 
	we design a centrality encoding that learns a centrality embedding based on the length of the \textit{influence duration} of each time step, which is formulated as
	\begin{equation}
	C_{i}^{t}=Linear\left( Deg\left( V_t \right),\varTheta _C \right); t \in \left[ 1,T \right],
	\end{equation} where $C_{i}^{t} \in \mathbb{R}^{D}$, $\varTheta _C$ is the set of learnable parameters, and $Deg\left( V_t \right)$ calculates the outdegree of the node $V_t$.
	The centrality encoding adaptively explores the importance of different time steps based on the influence duration. 

	For behavior-dependent temporal dependencies, we first adopt a non-linear multi-layer perceptron (MLP) to embed the position $X_i^t$ at each time step as
	\begin{equation}
	H_i^{t}=MLP \left( X_i^t,\varTheta _N \right),
	\end{equation} where $\varTheta _N$ contains the learnable parameters, and $H_i^{t} \in \mathbb{R}^{D}$.
	Besides, a positional embedding $P_{i}^t \in \mathbb{R}^{D}$ is applied to label the position of the motion state at each time step in a trajectory. In this paper, we encode the time step positions in a learnable way, as proposed in \cite{zhu2020deformable}.
	Then, we update the node embedding by
	\begin{equation}
	H_i^{t} := H_i^{t} + C_{i}^{t} + P_{i}^t.
	\end{equation} According to the Transformer encoder \cite{vaswani2017NIPS}, we further map $H_{i}$ into three values
	\begin{align}
	Q_i=H_{i}^{1:T}W_Q,\ K_i=H_{i}^{1:T}W_K,\ V_i=H_{i}^{1:T}W_V,
	\end{align}
	where $W_Q, W_K, W_V$ are the parameters corresponding to the Query $Q_i$, Key $K_i$ and Value $V_i$ of the pedestrian $i$. The output of temporal graphormer can be further computed as
	\begin{equation}
	\setlength{\abovedisplayskip}{3pt}
	\setlength{\belowdisplayskip}{3pt}
	Att\left( Q_i,K_i,V_i \right) =Softmax \left( \frac{Q_i\left( K_i \right) ^T} {\sqrt{d_k}} \odot A_{tmp} \right)V_i, 
	\label{attention}
	\end{equation} 
	where $A_{tmp}$ is the adjacency matrix, $\odot$ is the operation of dot product, and $d_k$ is the dimension of $Q_i$. Eq. (\ref{attention}) characterizes the behavior-dependent temporal dependencies, which builds the relationships among different motion behaviors across the temporal domain. 
	For brevity, we write the temporal graphormer as
	\begin{equation}
	TH_{i}^{1:T}=TG\left( X_{i}^{1:T}; \varTheta _{tg} \right),
	\end{equation}
	where  $TH_i^{1:T}$ is the \textit{Temporal Embedding} of the temporal graphormer, and $\varTheta _{tg}$ contains learnable parameters. By resorting to the temporal graphormer, both the behavior-independent temporal dependencies and behavior-dependent temporal dependencies can be characterized.
	\begin{figure*}[!t]
		\centering
		\includegraphics[width=0.70\textwidth]{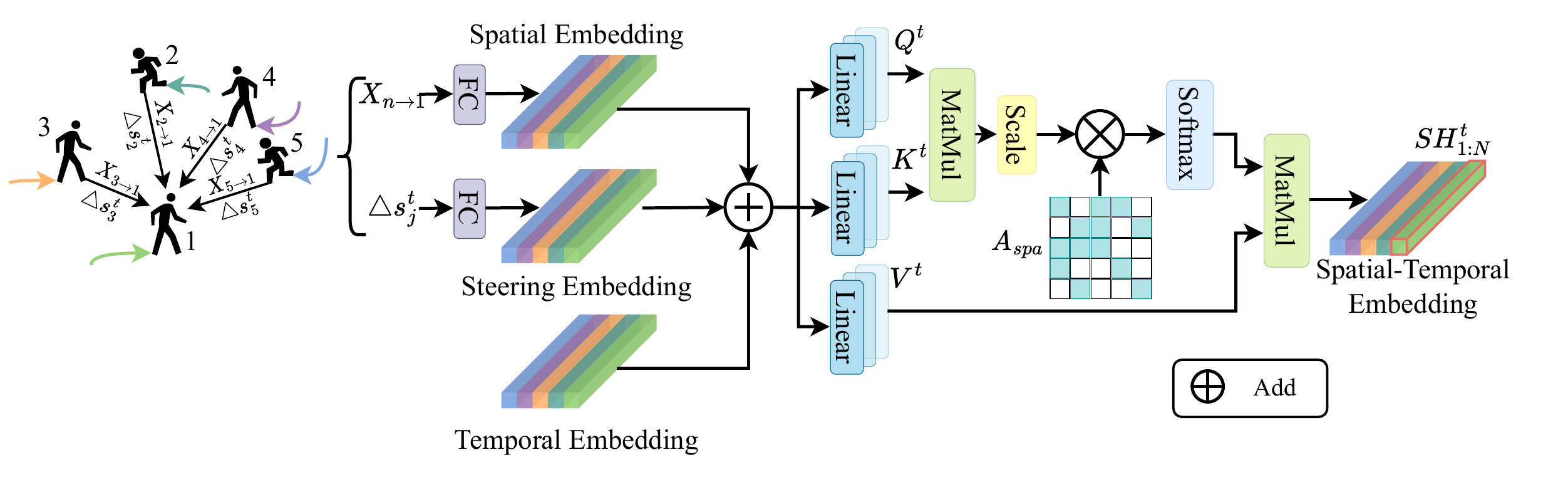}
		\caption{\centering Illustration of the spatial graphormer.}
		\label{fig:SG}
	\end{figure*}
	
	\subsubsection{Spatial Graphormer}
	Relative positions of other pedestrians are crucial for a target pedestrian to make decisions, such as adjusting velocity or direction to avoid collisions. 
	However, considering only the relative positions to the neighbors is not sufficient for making a reasonable decision. For example, when a neighbor in the rear is walking toward the opposite direction from the target pedestrian, the target pedestrian generally does not need to make adjustments, despite their close proximity. In reality, the relative steering angle of a neighbor to the target pedestrian and the walking direction of the target pedestrian are also key elements that drive pedestrians to change their motion states.
	
	
	To fully model the mutual spatial interactions between pedestrians, in this work, we propose a spatial graphormer by utilizing the relative motion states of the neighbors (\textit{i.e.}, the relative position and the relative steering angle to the target pedestrian) and the walking direction of the target pedestrian.
	Specifically, we take pedestrians in the scene as nodes and construct an undirected graph based on the walking direction of the target pedestrian and the relative motion states of the neighbors.
	First, at time step $t$, we can construct the spatial graph as
	\begin{equation}
	G_{spa}=(V_{spa}, A_{spa}),
	\end{equation}
	where $V_{spa}$ represents the nodes of $G_{spa}$, and $A_{spa}$ is the adjacency matrix of $G_{spa}$ describing the mutual spatial relationships among pedestrians. 
	To better capture spatial interactions, 
	we endow each node $V_{i}^{t}\in V_{spa}$ with the information of both relative positions and relative steering angles. Namely,
	\begin{align}
	V_{i}^{t}=R_{i}^{t}+S_{i}^{t}+TH_{i}^{t}, \label{eq:hidden_en}
	\end{align}
	where $TH_{i}^{t}$ denotes the temporal embedding of pedestrian $i$ at time step $t$, $R_{i}^{t}$ and $S_{i}^{t}$ represent the embedding of the relative position and relative steering angle of pedestrian $i$ to the target pedestrian, respectively. As shown in Fig. \ref{fig:SG}, we adopt a single layer MLP with ReLU activation to embed the relative position and relative steering angle of neighbors as
	\begin{equation}
	\begin{aligned}
	R_{j}^{t}=MLP\left( x_{j\rightarrow i}^{t}, y_{j\rightarrow i}^{t};\varTheta _P \right),\\
	x_{j\rightarrow i}^{t}=x_{j}^{t}-x_{i}^{t};\ y_{j\rightarrow i}^{t}=y_{j}^{t}-y_{i}^{t},
	\end{aligned}
	\end{equation}
	\begin{equation}
	\begin{aligned}
	S_{j}^{t}=&MLP ( \bigtriangleup s_{j}^{t};\varTheta _S ),\\
	\bigtriangleup s_{j}^{t}=&\frac{\bigtriangleup X_{i}^{t}\cdot \bigtriangleup X_{j}^{t}}{| \bigtriangleup X_{i}^{t} || \bigtriangleup X_{j}^{t} |},  
	\end{aligned}
	\end{equation}
	where $(x_{j\rightarrow i}^{t}, y_{j\rightarrow i}^{t})$ is the relative position of neighbor $j$ to the target pedestrian $i$ at time step $t$, $\bigtriangleup s_{j}^{t}$ denotes the corresponding relative steering angle, $\bigtriangleup X_{i}^{t}=(\bigtriangleup vx_{i}^{t}, \bigtriangleup vy_{i}^{t})=(x_{i}^{t}-x_{i}^{t-1}, y_{i}^{t}-y_{i}^{t-1})$ denotes the walking direction of pedestrian $i$ at time step $t$, and $\varTheta _P$, $\varTheta _S$ are the parameters of MLP. We refer to $R_{j}^{t}$ as the spatial embedding, and $S_{j}^{t}$ as the steering embedding. 
	Note that we do not employ positional encoding when modeling spatial interactions since there is no natural order for pedestrians in the scene.
	
	Intuitively, people commonly pay little attention to pedestrians outside their field of vision. For simplicity, we set the maximum binocular field of view to $180^\circ$ in the horizontal position for a pedestrian. Based on this fact, we design the adjacency matrix $A_{spa}$ as
	\begin{align}
	A_{spa}^{ij}=\left\{ \begin{array}{l}
	1,\ if\ x_{j\rightarrow i}^{t}\cdot \bigtriangleup vx_{i}^{t}\ge 0\ and\ y_{j\rightarrow i}^{t}\cdot \bigtriangleup vy_{i}^{t}\ge 0,\\
	-inf,\ \ \ \ \ \ \ \ \ \ \ \ \ \ \ \ \ \ Otherwise,\\
	\end{array} \right. \label{spatia A}
	\end{align}
	where $(\bigtriangleup vx_{i}^{t}, \bigtriangleup vy_{i}^{t})$ is the walking direction of pedestrian $i$ along $x$ and $y$ axes, and $(x_{j\rightarrow i}^{t}, y_{j\rightarrow i}^{t})$ is the relative position of neighbor $j$ to pedestrian $i$. $A_{spa}^{ij}$ characterizes the mutual spatial relationships among pedestrians.
	
	Similar to the temporal graphormer, the output of the spatial graphormer can be further computed as,
	\begin{equation}
	\begin{aligned}
	Q^t=SH_{1:N}^{t}W_{Q^{'}},\ K^t=SH_{1:N}^{t}W_{K^{'}},\ V^t=SH_{1:N}^{t}W_{V^{'}},\\
	Att\left( Q^t,K^t,V^t \right) =Softmax \left( \frac{Q^t\left( K^t \right) ^T}{\sqrt{d_k}}\odot A_{spa} \right)V^t, 
	\end{aligned}
	\end{equation}
	where $N$ is the number of pedestrians in the scene, $W_{Q^{'}}, W_{K^{'}}, W_{V^{'}}$ are the parameters. 
	For brevity, we write the spatial graphormer as
	\begin{equation}
	SH_{1:N}^{t}=SG\left( X_{1:N}^{t-1:t}, TH_{1:N}^{t}; \varTheta _{sh} \right),
	\end{equation}
	where $SH_{1:N}^{t}$ is the output of the spatial graphormer, which we define as \textbf{\textit{Spatial-Temporal Embedding}}. $\varTheta _{sh}$ is the set of parameters. 
	Similar to \cite{vaswani2017NIPS}, the multi-head attention mechanism is employed in our framework. 
	Note that different from graph-based methods, our dual graphormer takes the advantage of the self-attention mechanism to avoid direct aggregation of connected nodes, which could greatly alleviate the over-smoothing problem.
	Such a conclusion has been carefully justified in \cite{ying2021transformers}.
	
	\subsubsection{Proposed Dual Graphormer in Trajectory Prediction}
	In order to model the behavior pattern of pedestrians, we first apply the Dual Graphormer to extract the deep representations of \textbf{\textit{motion behaviors}} and \textbf{\textit{social interactions}}. As shown in Fig. \ref{fig:framework}, we primarily extract representations of two types of inputs, \textit{i.e.}, the full trajectory and the observed historical trajectory. Note that the full trajectories are used only in the training phase.

	The full trajectory reflects the motion behavior of each pedestrian in a time period, such as where to go and how to go. In our STGlow model, we define the representation of the full trajectory as the \textbf{\textit{motion behavior}} of each pedestrian, which can be formulated as
	\begin{equation}
	MB_{target}=TG\left( Concat\left( X_{target}, Y_{target} \right), \varTheta _{tg} \right),
	\end{equation}
	where $TG(\cdot)$ is the temporal graphormer.

	Besides, the observed historical trajectories of all pedestrians in the scene are encoded via the temporal and spatial graphormer to extract the \textbf{\textit{social interactions}} of the target pedestrian, which can be written as
	\begin{equation}
	\begin{aligned}
	TH_{1:N}=TG\left( X_{1:N}; \varTheta _{tgh} \right),\\
	SH_{target}=SG\left( X_{1:N}^{-1:0}, TH_{1:N}^{0}; \varTheta _{sgh}\right),\\
	TH_{target}=TG\left( X_{target}; \varTheta _{tgy} \right),\\
	ST_{target}=TH_{target} + SH_{target}.
	\end{aligned}  
	\end{equation}
	where $\varTheta _{tgh}$, $\varTheta _{sgh}$ and $\varTheta _{tgy}$ are parameters. 
	So far, we have obtained the representation $MB_{target}$ of \textbf{\textit{motion behavior}} of the target pedestrian and the mutual \textbf{\textit{social interaction}} $ST_{target}$.
	In the inference phase, we first predict the latent motion behavior $MB_{target}$ conditioned on the $ST_{target}$. The $MB_{target}$ is then fed into a decoder to generate future trajectories.

	\subsection{Proposed Generative Flow with Pattern Normalization (Glow-PN)}\label{sec:glow}
	Due to the differences of human intent and unique behavior patterns, the motion behaviors of pedestrians are of high diversity.
	Namely, the future trajectories could be very different given the same historical trajectory. 
	Thus, learning the underlying distribution of motion behaviors conditioned on social interactions is important to predict the diverse trajectories of pedestrians. 
	Existing approaches commonly use GANs or CVAEs to generate diverse trajectories, where their limitations have been carefully discussed in the first section. 

	\begin{figure}[!t]
		\centering
		\includegraphics[width = 0.25\textwidth]{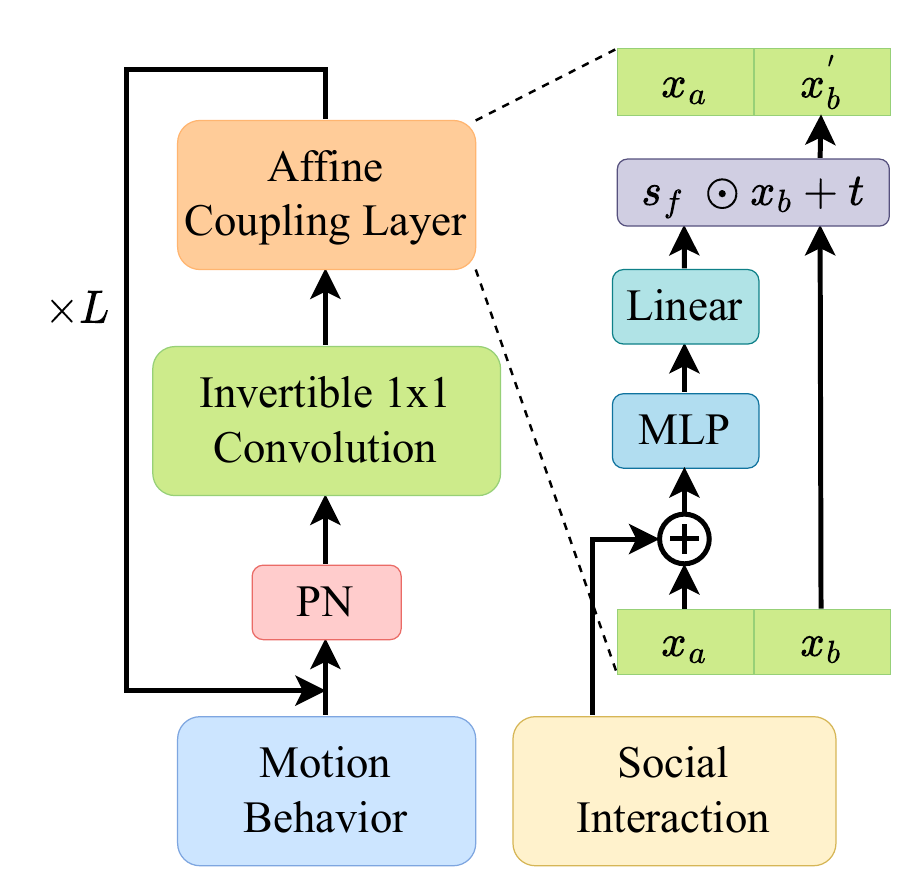}
		\caption{Illustration of the proposed generative flow with Pattern Normalization (Glow-PN).}
		\label{fig:Glow}
	\end{figure}
	In our work, we simulate the process of degradation and evolution between the complex motion behavior and a simple behavior as follows:
	\begin{align}
	MB_{target} \stackrel{f_1}{\longleftrightarrow}h_1&\stackrel{f_2}{\longleftrightarrow}h_2\cdots \stackrel{f_k}{\longleftrightarrow}z,
	\label{MB_z} 
	\end{align}
	where $\{f_1, f_2,...,f_k\}$ denote a set of \textit{invertible} transformations to simulate the degrading and evolving process, and $h_1, ...,h_{k-1}$ denote the intermediate motion behaviors. We can see that the forward process of formula (\ref{MB_z}) gradually degrades the complex motion behavior $MB_{target}$ into a simple behavior $z$, while its reverse process represents the evolution of a simple behavior to the complex motion behavior.

	In reality, the complex motion behavior $MB_{target}$ of pedestrians is not accomplished at one stroke, but based on simple motion behaviors $z$ (\textit{e.g.}, every step of walking or every action) combined with individual behavior habits which may originate from various aspects such as people's walking habits, traffic awareness, and potential intentions.
	For the sake of simplicity, we assume that $z$ follows a standard normal distribution, \textit{i.e.}, 
	\begin{equation}
	\begin{aligned}
	z\thicksim \mathcal{N}&\left( z;0,I \right), \\
	\end{aligned}
	\label{MB}
	\end{equation}
	where $I$ is an identity matrix. Letting $x \triangleq MB_{target}$, the log-likelihood of complex motion behavior can be written as:
	\begin{align}
	\log\text{\ }p_{\theta}\left( x \right) &= \log \left( p_{\theta}\left( z \right) |\det \left( dz/dx \right) | \right) \label{first} \\ 
	&=\log p_{\theta}\left( z \right) \ +\ \sum_{i=1}^K{\log \det \left( dh_i/dh_{i-1} \right) |} \label{second} \\ 
	&= \log p_{\theta}\left( z \right) +\sum_{i=1}^K{\log \det \left( J\left( f_{i}^{-1}(x) \right) \right) |}. \label{log}
	\end{align} 
	The equation (\ref{first}) holds because $\int_x{p_{\theta}\left( x \right) dx}=\int_z{p_{\theta}\left( z \right) dz}$, and the equation (\ref{log}) holds because $h_i=f_i^{-1}\left( h_{i-1} \right)$.
	The first term in formula (\ref{log}) is the log-likelihood of the standard normal distribution and the scalar value ${\log{|}\det \left( J\left( f_{i}^{-1}\left( x \right) \right) \right) |}$ is the logarithm of the absolute value of the determinant of the Jacobian matrix $J\left( f_{i}^{-1}\left( x \right) \right)$. This value reflects the transformation from $h_{i-1}$ to $h_i$ of the motion behavior under the transformation $f_i$. Then, our framework optimizes the parameters by minimizing the negative log-likelihood function, where the loss is 
	\begin{equation}\label{eq:LP}
	L_p=\min~-\sum_{i=1}^N{\log~p_{\theta}\left( x_i \right)}.
	\end{equation} In contrast to previous approaches, our method can more precisely model the underlying data distribution by optimizing the exact log-likelihood of motion behaviors as shown in (\ref{eq:LP}). Obviously, how to design the invertible transformation functions $f_1,...,f_k$ is essential for the optimization of (\ref{eq:LP}). We should bear in mind that $f_1,...,f_k$ should be differentiable thus allowing the end-to-end training.

	In our work, we employ Glow \cite{kingma2018glow} to learn the invertible transformations of motion behaviors from complex to simple. As shown in Fig. \ref{fig:Glow}, for the forward process, we take \textbf{\textit{motion behavior}} $MB_{target}$ as input of Glow-PN. After several ``steps of flow", the complex \textbf{\textit{motion behavior}} is degraded into a simple behavior that can be represented by a simple distribution. A step of flow here consists of a pattern normalization (PN), an invertible $1 \times 1$ convolution, and an affine coupling layer, described below.

	\subsubsection{Pattern Normalization (PN)} 
\begin{figure}[!t]
		\centering
		\includegraphics[width = 0.30\textwidth]{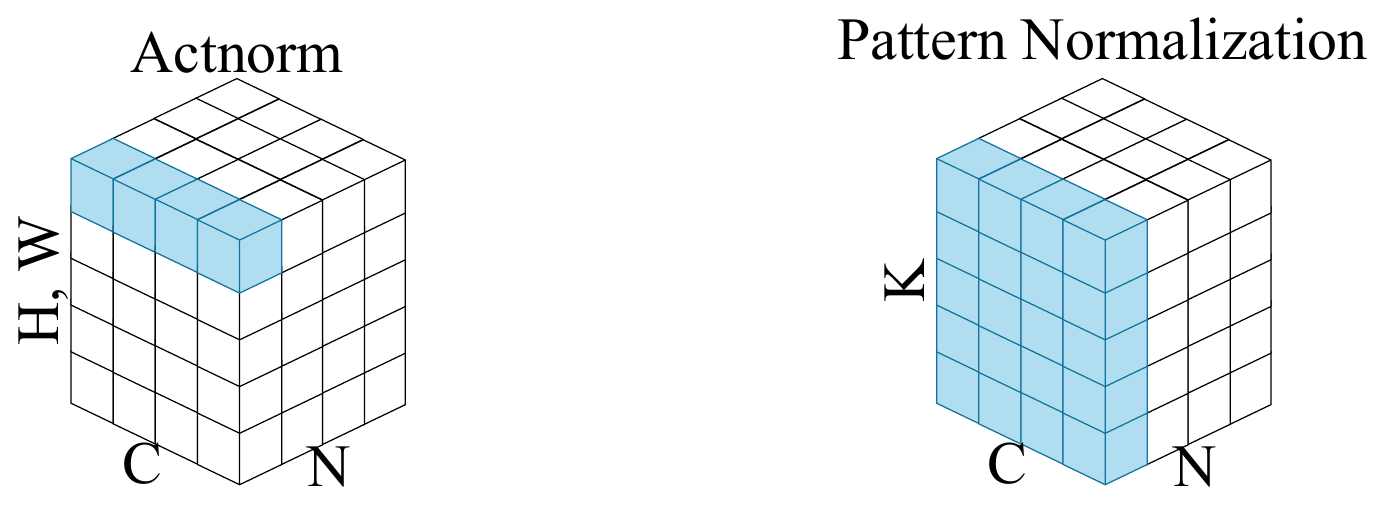}
		\caption{In ActNorm, (H, W) as the size of the feature map and C as the channel axis, while in PN, K as the sample axis and C as the feature axis. N as the batch axis. The pixels in blue are normalized by the same learnable scales and bias.}
		\label{fig:PN}
	\end{figure}

	Considering the fact that diverse motion behaviors share the same behavior pattern (\textit{e.g.}, the individual walking habits), we propose a pattern normalization (PN) for trajectory prediction as shown in Fig. \ref{fig:PN}. Different from the Actnorm adopted in the original Glow, PN performs normalization for multiple motion behaviors jointly at the sample axis. As will be shown in the experiments, our proposed PN greatly improves the prediction performance.

	Specifically, the forward function, reverse function and log-determinant of our proposed PN are calculated as:
	\begin{equation}
	\begin{aligned}
	Function:&\ y=s_g\odot x+b, \\
	Reverse\ Function:&\ x=\left( y-b \right) /s_g, \\
	Log\ determinant:&\ \log |s_g|,
	\end{aligned}
	\end{equation}
	where $x$ indicates the input of PN, and $y$ signifies its output. Both $x$ and $y$ are tensors of shape $\left(K\times C\times 1\right)$ with $K$ motion behaviors and channel dimension $C$. $s_g,b \in \mathbb{R}^{K \times C}$ are learnable scales and bias parameters whose initialization depends on the input data, so that the data has zero mean and unit variance after PN. After initialization, the scale and bias are treated as regular trainable parameters that are independent of the data. 
	Note that, different from Layer Normalization \cite{ba2016layer}, to ensure that the transformation $f$ is invertible, we design the forward and reverse process of normalization in PN. The forward process normalizes the behavior pattern of each pedestrian to promote model convergence, while the reverse process de-normalizes multiple motion behaviors to restore its original motion characteristics.
	\subsubsection{Invertible $1\times 1$ Convolution}
	In order that all channels of input in the forward transformation can be updated in subsequent coupling layers, we following Glow apply an invertible $1 \times 1$ convolution layer before coupling layers.
	The weights $W$ of convolution are initialized as a random rotation matrix and hence invertible. Thus, the log-determinant of this transformation is easy to compute:
	\begin{align}
	f_{conv}^{-1}&=Wy, \\
	\log |\det \left( J\left( f_{conv}^{-1}\left( y \right) \right) \right) |&=\log |\det \left( W \right) |,
	\end{align}
	where the log-determinant starts at zero and after one SGD step, the values start to diverge from zero.
	\subsubsection{Affine Coupling Layer}
	Generally, computing the determinants of high-dimensional Jacobian and large matrices is very expensive. Following Glow, we reduce the complexity by designing tractable and flexible invertible transformation. Specifically, we introduce an affine coupling layer, which can efficiently compute forward function, reverse function and log-determinant.
	\begin{align}
	x_a,x_b&=split\left( x_{affine} \right),\\
	\left( \log s_f, t \right) =Linear&\left( MLP\left( concat\left( x_a,ST_{target} \right) \right) \right), \label{NN}\\
	x_{b}^{'}&=s_f\odot x_b+t, \\
	x^{'}=c&oncat\left( x_a,x_{b}^{'} \right),
	\end{align}
	where $x_{affine} \in \mathbb{R}^{1\times C\times K}$ represents the input of affine coupling layer, the $split\left(\cdot\right)$ function splits $x$ into two havles along the channel dimension, while the $concat\left(\cdot\right)$ operation performs the restore operation. Since the inputs $x_a$ of formula (\ref{NN}) remains unchanged during the affine transformation operation, formula (\ref{NN}) can be arbitrary transformation. Accordingly, in the reverse process of the affine coupling layer, $s_f$ and $t$ can be obtained from the output $x_a$ through formula (\ref{NN}), and $x_{b}^{'}$ can be obtained by the reverse of $x_b$. Note that our affine coupling layer is conditioned on the social interaction $ST_{target}$ to establish the mapping of complex to simple motion behaviors.

	As we can see, only the $s_f$ term changes the volume of the mapping in the affine coupling layer and adds a change of variables term to the loss. The log-determinant of the affine coupling layer and the final likelihood are hence computed as follows:
	\begin{equation}
	\log |\det \left( J\left( f_{coupling}^{-1}\left( x \right) \right) \right) |=\log |s_f|.
	\end{equation}
	\begin{equation}
	\begin{split}
	\log p_{\theta}\left( x \right) &=\log p_{\theta}\left( z \right) \\
	&+\sum_{j=0}^n \left( \log |s_{g}^{j}|+\log \det |W^j|+\log |s_{f}^{j}| \right),
	\end{split}
	\end{equation}
	where $p_{\theta}\left( z \right) =\mathcal{N}\left( z;0,I \right)$, and its log-likelihood can be easily computed as: $-z\left( x \right) ^Tz\left( x \right) /2\sigma ^2$, where $\sigma=I$. So far, we have eventually completed the design of the invertible transformation of ``step of flow". Note that, different from CVAE-based methods, our method optimizes the exact log-likelihood of motion behaviors.

	\begin{figure}[!t]
		\centering
		\includegraphics[width = 0.30\textwidth]{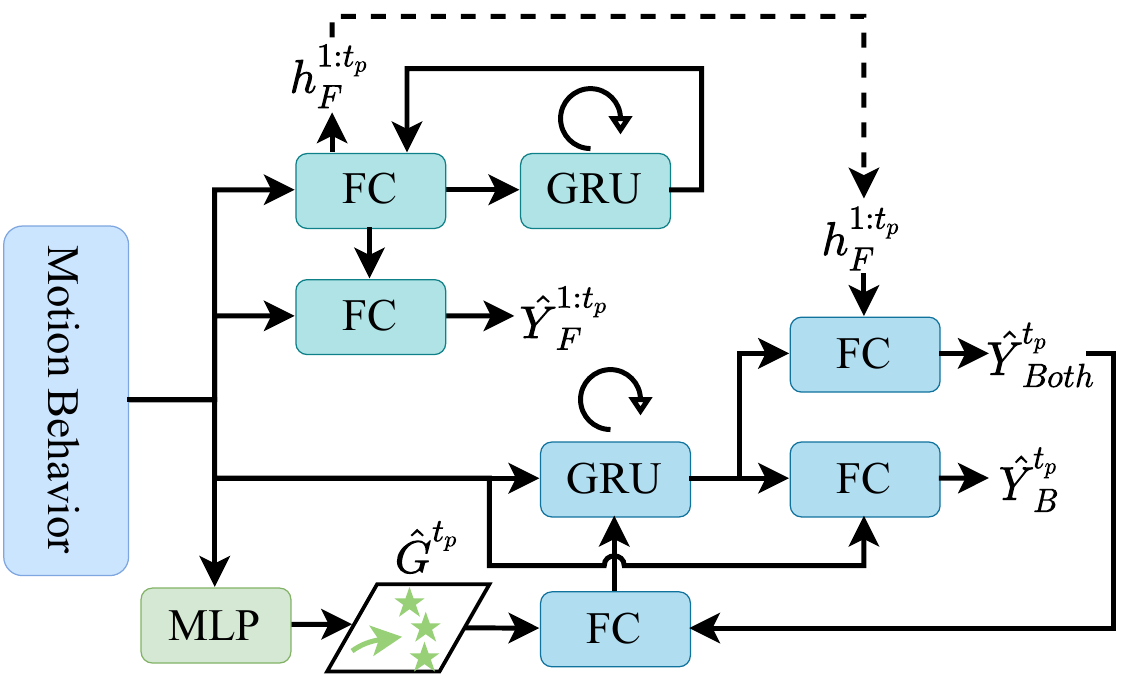}
		\caption{\centering Illustration of the bi-directional trajectory prediction module.}
		\label{fig:Bi-Trajectory}
	\end{figure}
	\subsection{Bi-directional Trajectory Prediction}
	Due to the invertible design of Glow-PN, a simple behavior drawn from the standard normal distribution can be directly evolved into a complex motion behavior $MB_{target}$ through the reverse process of Glow-PN, conditioned on the social interaction $ST_{target}$. Then, we feed the obtained $MB_{target}$ into the decoder to predict the future trajectory.

	\subsubsection{Bi-directional Decoder with Goal Estimation}
	To alleviate the error accumulation issue caused by the recurrent neural network in trajectory prediction, as shown in Fig. \ref{fig:Bi-Trajectory}, we adopt the decoder proposed in Bitrap \cite{yao2021bitrap}, which is designed in a bi-directional manner.
	We additionally predict and supervise the forward and backward future trajectory to strengthen the representation learning during bidirectional prediction.
	Specifically, we first predict the goal of each pedestrian as
	\begin{equation}
	\hat{G}^{t_p}=MLP\left( MB_{target}^{'};\varTheta _g \right),
	\end{equation}
	where $MB_{target}^{'}$ is the representation of complex motion behaviors evolved by the reverse process of Glow-PN from simple behaviors. Then, the forward trajectory $\hat{Y}_{F}^{t}$ is predicted as 
	\begin{equation}
	\begin{aligned}
	f_{i}^{t-1}&=MLP\left( f_{h}^{t-1};\varTheta _{fi} \right),\\
	f_{h}^{t}&=GRU\left( f_{h}^{t-1},f_{i}^{t-1};\varTheta _{ff} \right),\\
	\hat{Y}_{F}^{t}&=FC\left( concat\left( f_{i}^{t},MB_{target}^{'} \right) ;\varTheta _{fy} \right),    
	\end{aligned}
	\end{equation}
	where $t$ from $1$ to $t_p$ represents the time step of the forward prediction, and $f_{h}^{0}=MLP\left( MB_{target}^{'};\varTheta _{fh} \right)$. Further, the backward prediction $\hat{Y}_{B}^{t_b}$ is formulated as
	\begin{equation}
	\begin{aligned}
	b_{h}^{t_b}&=GRU\left( b_{i}^{t_b+1},b_{h}^{t_b+1};\varTheta _{bh} \right),\\
	\hat{Y}_{B}^{t_b}&=FC\left( concat\left( b_{h}^{t_b},MB_{target}^{'} \right) ;\varTheta _{by} \right),
	\end{aligned}
	\end{equation}
	where $t_b$ from $t_p-1$ to $1$ denotes the time step of the backward prediction, and $b_{h}^{t_p}=MLP\left( MB_{target}^{'};\varTheta _{bh} \right)$. Last, the bidirectional prediction $\hat{Y}_{Both}^{t_b}$ considering both forward and backward trajectories is predicted as
	\begin{equation}
	\begin{aligned}
	\hat{Y}_{Both}^{t_b}&=FC\left( concat\left( b_{h}^{t_b},f_{i}^{t_b} \right) ;\varTheta _{both} \right),\\
	b_{i}^{t_b}&=MLP\left( \hat{Y}_{Both}^{t_b},\varTheta _{bi} \right),
	\end{aligned}
	\end{equation}
	where $b_{i}^{t_p}=MLP\left( \hat{G}^{t_p},\varTheta _{bi} \right)$. Besides, $\varTheta_{*}$ indicates the set of parameters. We supervise the predicted forward, backward and bidirectional trajectories as
	\begin{equation}
	\begin{aligned}
	L_{traj} &= \underset{k\in K}{\min}~\alpha \lVert G_{k}^{t_p}-\hat{G}_{k}^{t_p} \rVert +\underset{k\in K}{\min}\sum_{t = 1}^{t_p}{( \lambda _1\lVert Y_{k}^{t}-\hat{Y}_{F}^{t} \rVert }\\
	& +\lambda _2\lVert Y_{k}^{t}-\hat{Y}_{B}^{t} \rVert +\lambda _3\lVert Y_{k}^{t}-\hat{Y}_{Both}^{t} \rVert ),
	\end{aligned}
	\end{equation}
	where $K$ represents the number of samples of future trajectories, and $\alpha$, $\lambda _1$, $\lambda _2$, $\lambda _3$ are the coefficients that balance different losses. Combined with the log-likelihood of the motion behaviors, our final loss is formulated as
	\begin{align} \label{eq:final_loss}
	L_{total}=L_p + \sum_{i=1}^N{L_{traj}}.
	\end{align}
	
	\subsubsection{Inference}
	In the inference phase, we can easily generate the diverse trajectories by simple behaviors sampled from the standard normal distribution through the reverse process of Glow-PN conditioned on social interactions.
	\begin{table*}[!t]
		\renewcommand{\arraystretch}{1.0}
		\centering  
		\caption{Quantitative results on the ETH/UCY dataset with Best-of-20 strategy in ADE/FDE metric. Lower is better.}
		\label{tb:comparison}  
		\begin{tabular}{c|c|c@{\hspace{1mm}}c|c@{\hspace{1mm}}c|c@{\hspace{1mm}}c|c@{\hspace{1mm}}c|c@{\hspace{1mm}}c|c@{\hspace{1mm}}c}  
			\hline  
			\multirow{2}{*}{\diagbox{\textbf{Methods}}{\textbf{Datasets}}}&\multirow{2}{*}{\textbf{Sampling}}&\multicolumn{2}{c|}{\textbf{ETH}}&\multicolumn{2}{c|}{\textbf{Hotel}}&\multicolumn{2}{c|}{\textbf{Univ}}&\multicolumn{2}{c|}{\textbf{Zara1}}&\multicolumn{2}{c|}{\textbf{Zara2}}&\multicolumn{2}{c}{\textbf{AVG}}\cr\cline{3-14}&&\textbf{ADE}&\textbf{FDE}&\textbf{ADE}&\textbf{FDE}&\textbf{ADE}&\textbf{FDE}&\textbf{ADE}&\textbf{FDE}&\textbf{ADE}&\textbf{FDE}&\textbf{ADE}&\textbf{FDE}\cr
			\hline
			S-GAN\cite{gupta2018social}&20&0.87&1.62&0.67&1.37&0.76&1.52&0.35&0.68&0.42&0.84&0.61&1.21\cr
			Social-STGCNN\cite{mohamed2020social}&20&0.64&1.11&0.49&0.85&0.44&0.79&0.34&0.53&0.30&0.48&0.44&0.75\cr 
			GTPPO \cite{9447207} &20 &0.63&0.98&0.19&0.30&0.35&0.60&0.20&0.32&0.18&0.31&0.31&0.50\cr
			TPNMS\cite{liang2021temporal}&20&0.52&0.89&0.22&0.39&0.55&1.13&0.35&0.70&0.27&0.56&0.38& 0.73\cr
			TPNSTA\cite{Li2022}&20&0.51&0.87&0.22&0.39&0.52&1.09&0.34&0.68&0.26&0.54&0.37&0.71\cr
			STAR\cite{yu2020spatio}&20&0.56&1.11 &0.26&0.50& 0.52&1.15 &0.41&0.90& 0.31&0.71& 0.41&0.87\cr
			Trajectron++\cite{salzmann2020trajectron++}&20& 0.39& 0.83& \color{red}{\underline{0.12}}& \color{red}{\underline{0.21}}& 0.20& 0.44& 0.15& 0.33& 0.11& 0.25& 0.19& 0.41\cr
			AgentFormer\cite{yuan2021agentformer}&20& 0.45&0.75& 0.14&0.22& 0.25&0.45& 0.18&0.30& 0.14&0.24& 0.23&0.39\cr
			PECNet\cite{mangalam2020not}&20& 0.54&0.87& 0.18&0.24 &0.35&0.60 &0.22&0.39& 0.17&0.30& 0.29&0.48\cr
			SGCN\cite{shi2021sgcn}&20&0.63&1.03& 0.32&0.55& 0.37&0.70& 0.29&0.53& 0.25&0.45& 0.37&0.65\cr 
			DMRGCN\cite{bae2021disentangled}&20&0.60&1.09& 0.21&0.30& 0.35&0.63& 0.29&0.47& 0.25&0.41& 0.34&0.58\cr
			Y-Net+TTST\cite{mangalam2021goals}&\underline{\bf10000}& \underline{\bf0.28}& \underline{\bf0.33}& 0.10& \bf0.14& 0.24& 0.41& 0.17& 0.27& 0.13& 0.22& 0.18& 0.27\cr
			CAGN\cite{duan2022complementary}& 20& 0.41& \color{red}{\underline{0.65}} &0.13&0.23& 0.32&0.54& 0.21&0.38& 0.16&0.33& 0.25&0.43\cr
			MID\cite{gu2022stochastic}&20& 0.39& 0.66 &0.13& 0.22& 0.22& 0.45& 0.17& 0.30& 0.13& 0.27& 0.21& 0.38\cr
			BiTraP\cite{yao2021bitrap}&20&\color{red}{\underline{0.37}}&0.69&\color{red}{\underline{0.12}}&\color{red}{\underline{0.21}}&\color{red}{\underline{0.17}}&\color{red}{\underline{0.37}}&\color{red}{\underline{0.13}}&\color{red}{\underline{0.29}}&\color{red}{\underline{0.10}}&\color{red}{\underline{0.21}}&0.18&0.35\cr
			\bf STGlow (ours)&20&\bf0.31&\bf0.49&\bf0.09&\bf0.14&\bf0.16&\bf0.33&\bf0.12&\bf0.24&\bf0.09&\bf0.19&\bf0.15&\bf0.28\cr
			\hline
			Improvement & 20 & \bf16\% & \bf25\% & \bf25\% & \bf33\% & \bf6\% & \bf11\% & \bf8\% & \bf17\% & \bf10\% & \bf10\% & \bf13\% & \bf19\% \cr
			\hline
		\end{tabular}
	\end{table*}
	\section{Experiments}  \label{sec:experiment}
	In this section, we evaluate the performance of our proposed STGlow, which is implemented using the PyTorch framework. All the
	experiments are performed on Ubuntu 18.04 with an NVIDIA 3090 GPU. Our source code and trained models will be publicly available upon acceptance. 
	
	\subsection{Implementation Details}
	The dimension of the node embedding is set as $256$, and the number of coupling layers and invertible $1 \times 1$ convolutions are empirically set as $16$. We also output $64$ of the channels after every $4$ coupling layers. In the training phase, the number of samples of predicted trajectories is set to $20$ and the coefficients of the final loss are set to be $1$, $0.25$, $0.25$, $0.5$, respectively. We adopt Adam algorithm \cite{kingma2014adam} to optimize the loss function (\ref{eq:final_loss}) and train our network with the following hyper-parameter settings: batch size is $128$; learning rate is 1e-3; betas are 0.9 and 0.999; weight decay is 1e-6 and the number of epochs is 400. 
	\subsection{Datasets and Metrics}
	\textit{Datasets}: 
	We evaluate our method on two benchmark public pedestrian trajectory prediction benchmarks including ETH/UCY dataset \cite{pellegrini2009you, lerner2007crowds} and Stanford Drone Dataset (SDD)\cite{robicquet2016learning}.
	The ETH and UCY dataset group consists of $5$ video sequences: ETH \& HOTEL (from ETH) and UNIV, ZARA1, \& ZARA2 (from UCY). All trajectories are converted to world coordinates, so the results we report are in meters.
	SDD comprise of more than $11000$ unique pedestrians across 20 top-down scenes captured on the stanford university campus in bird’s eye view containing several moving agents like humans and vehicles. We use the standard test train split as used in \cite{mangalam2020not} and other previous works.
	

	\textit{Metrics}: For the sake of fairness, we adopt the evaluation metrics Average Displacement Error (ADE) and Final Displacement Error (FDE) which are commonly used in literature\cite{alahi2016social,gupta2018social,yuan2021agentformer,mohamed2020social}. ADE computes the average $\ell 2$ distance between the predictions and the ground truth future while FDE computes the $\ell 2$ distance between the predicted and ground truth at the last observed point. The number of observed time steps is $8$ ($3.2$ seconds) of each person and the upcoming trajectory of $12$ time steps ($4.8$ seconds) is used to predict. For the ETH/UCY dataset, we use the widely adopted leave-one-out approach evaluation methodology such that we train our model on four scenes and test on the remaining one\cite{gupta2018social,liang2021temporal,Li2022}. Considering the diversity of future trajectories, we use the Best-of-K strategy to compute the final ADE and FDE with $K = 20$. 
	

	\subsection{Baselines}
	We compare with the following baselines including previous state-of-the-art methods:\\ 
	\textit{\textbf{GAN-based methods:}}
	S-GAN \cite{gupta2018social}: a model that employs GAN with a global pooling module to generate diverse pedestrian trajectories; 
	TPNMS \cite{liang2021temporal} and TPNSTA \cite{Li2022}: methods based on the temporal pyramid network to model global and local context of motion behavior, the latter further designs the spatial-temporal attention mechanism.\\
	\textit{\textbf{CVAE-based methods:}} 
	Trajectron++ \cite{salzmann2020trajectron++}: a recurrent graph based forecasting model incorporating dynamic constrains; 
	PECNet \cite{mangalam2020not}: a goal conditioned trajectory prediction network;
	BiTraP \cite{yao2021bitrap}: a goal-conditioned bidirectional trajectory prediction method based on the CVAEs.\\
	\textit{\textbf{Graph-based methods:}} 
	Social-STGCNN \cite{mohamed2020social}: an approach that models the social behavior of pedestrians using a graph;
	SGCN \cite{shi2021sgcn}: an approach that models the sparse directed interaction with a sparse directed spatial graph;
	DMRGCN \cite{bae2021disentangled}: a model that introduce a disentangled multi-scale aggregation to represent social interactions;
	GTPPO \cite{9447207}: a Graph-based Trajectory Predictor with Pseudo-Oracle.\\
	\textit{\textbf{Transformer-based methods:}}
	AgentFormer \cite{yuan2021agentformer}: a transformer-based approach that models the time and spatial dimensions simultaneously;
	STAR \cite{yu2020spatio}: a spatial-temporal graph transformer framework.\\
	\textit{\textbf{Other methods:}} 
	Y-Net \cite{mangalam2021goals}: method based on position and visual image information. `TTST' stands for the Test-Time Sampling Trick (TTST) in post-processing, which first samples 10000 trajectories and then clusters them into 20 trajectories;
	CAGN \cite{duan2022complementary}: a complementary attention gated network for pedestrian trajectory prediction;
	SIT \cite{shi2022social}: a tree-based method for pedestrian trajectory prediction;
	MID \cite{gu2022stochastic}: a method based on the diffusion model.

	\subsection{Quantitative Analysis}
	We quantitatively compare our \textbf{\textit{STGlow}} with a wide range of current methods. Table \ref{tb:comparison} compares our method with existing algorithms on the ETH/UCY dataset. Besides the performance on each dataset, we report the average results for each method in the last two columns. 
	Noted that, we also report the sampling number since adding the sampling number can effectively promote the performance \cite{gu2022stochastic}. 
	Based on the results, we draw the following conclusions:
	
	\begin{itemize}
		\item In general, with the same sampling number of 20, our method STGlow outperforms all the previous approaches in terms of ADE and FDE for all datasets. The last row of Table \ref{tb:comparison} shows the performance improvement of our method over the previous best methods (marked with red underline), where our method improves the ADE/FDE metrics by an average of 13\%/19\% on ETH/UCY datasets, respectively.
		\item Compared with the GAN-based method \cite{gupta2018social,liang2021temporal,Li2022}, our method achieves significant performance gains on ADE and FDE metrics. For example, Our method achieves $59\%$ and $61\%$ relative improvements in average ADE and FDE metrics over the GAN-based method TPNSTA. In addition, compared with the methods based on CVAEs\cite{salzmann2020trajectron++,yuan2021agentformer,mangalam2020not,yao2021bitrap} and diffusion model\cite{gu2022stochastic}, our method also achieves a greater performance improvement due to the optimization of the exact log-likelihood of motion behavior rather than the variational lower bound. For instance, our method improves the average ADE and FDE metrics by $17\%$ and $20\%$ respectively compared with the best CVAE-based method BiTraP\cite{yao2021bitrap}, and $29\%$ and $26\%$ respectively compared with MID\cite{gu2022stochastic}.
		\item Compared with the graph-based approach\cite{9447207,mohamed2020social,bae2021disentangled,shi2021sgcn} that models social interactions in an intuitive way, we integrate the graph structure with the Transformer structure in modeling social interactions and achieve significant performance improvements. For example, compared with DMRGCN\cite{bae2021disentangled}, our STGlow improves $56\%$ and $52\%$ in average ADE and FDE metrics respectively.
		\item Despite the unfair experimental settings, our method still achieves 17\% and 4\% performance improvement over Y-Net+TTST on average ADE/FDE metrics respectively.
	\end{itemize}
	\begin{table}[!t]
		\renewcommand{\arraystretch}{1.0}
		\centering  
		\caption{Quantitative results on the SDD dataset with Best-of-20 strategy in ADE/FDE metric. $*$ means the results are reproduced by \cite{gu2022stochastic} with the official released code. Lower is better.}
		\label{tb:SDD}  
		\begin{tabular}{c|c|c|c}
			\hline
			\textbf{Methods} & \textbf{Sampling} & \textbf{ADE}  & \textbf{FDE}   \\ \hline
			S-GAN\cite{gupta2018social}       & 20                & 27.23         & 41.44          \\
			PECNet\cite{mangalam2020not}           & 20                & 9.96          & 15.88          \\
			Y-Net + TTST\cite{mangalam2021goals}     & 10000             & 7.85          & 11.85          \\
			Y-Net$^{*}$\cite{mangalam2021goals}         & 20                & 8.97          & 14.61          \\
			GTPPO \cite{9447207} & 20                & 10.13          & 15.35          \\
			Trajectron++$^{*}$\cite{salzmann2020trajectron++}  & 20                & 8.98          & 19.02          \\
			SIT\cite{shi2022social}                     & 20          & 8.59              &    15.27    \\
			MID\cite{gu2022stochastic}              & 20                & 7.61          & 14.30          \\
			\textbf{STGlow (ours)}    & 20                & \textbf{7.20} & \textbf{11.20} \\ \hline
		\end{tabular}
	\end{table}
	
	We can draw similar conclusions above on SDD dataset, as shown in Table \ref{tb:SDD}, which indicates that the proposed STGlow has better generalization performance. 
	It is worth noting that STGlow achieved a relative improvement of $20\%$ and $23\%$ in ADE and FDE metrics compared to Y-Net\cite{mangalam2021goals} without using `TTST' post-processing.
	Furthermore, compared with the current state-of-the-art method MID\cite{gu2022stochastic}, our method achieves the best performance on ADE and FDE metrics.
	\begin{figure*}[!t]
		\centering
		\includegraphics[width=0.80\textwidth]{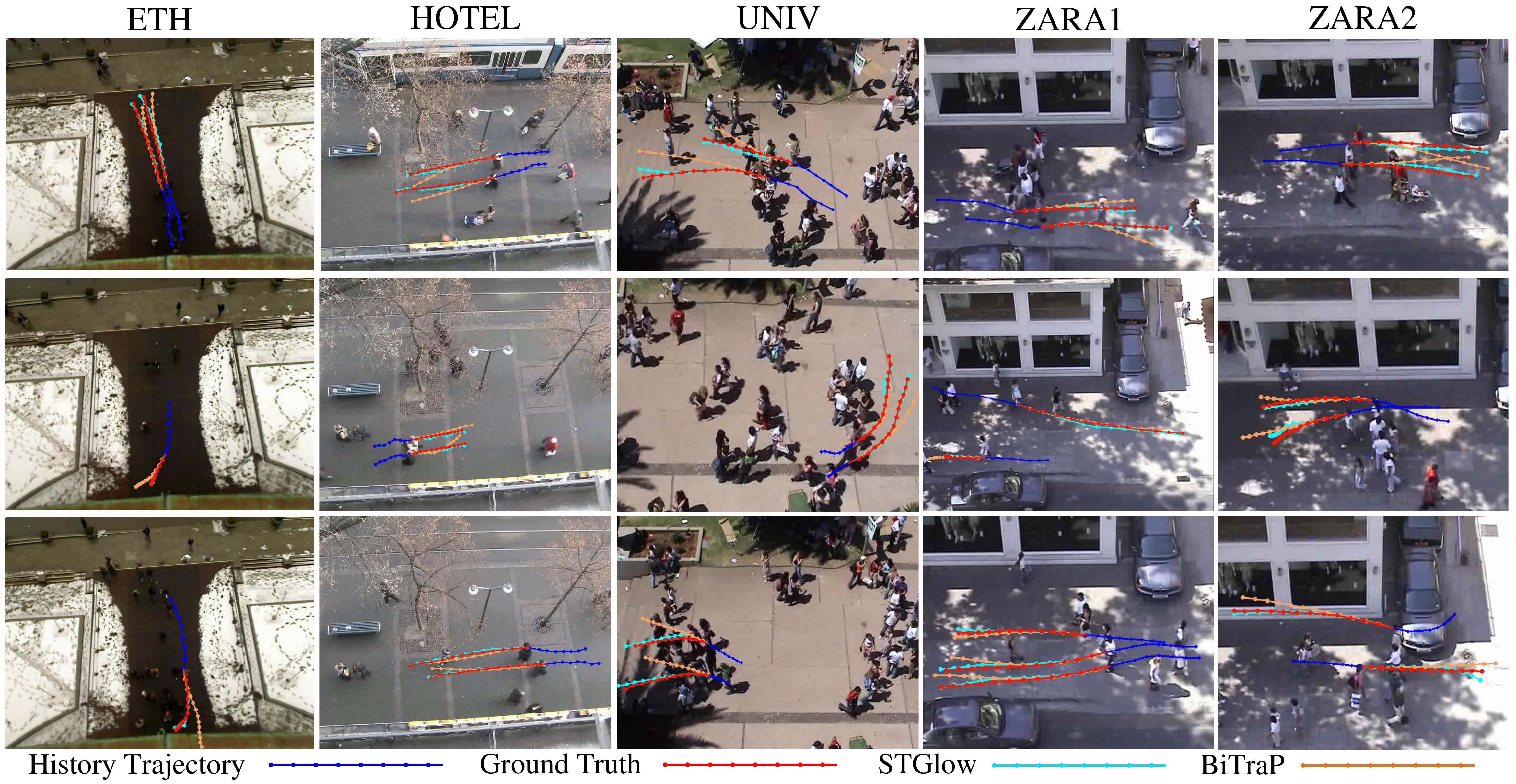}
		\caption{Visualization of predicted trajectories on the ETH/UCY Dataset. Given the history trajectories (blue line), we illustrate the ground truth paths (red line) and predicted future trajectories by our STGlow (green line) and the previous state-of-the-art BiTrap (orange line) for five different scenes, and dots are the locations of pedestrians at different time steps. Best viewed in color and zoom-in for more clarity.}
		\label{fig:qual_result}
	\end{figure*}
	\begin{figure}[!t]
		\centering
		\includegraphics[width=1.0\linewidth]{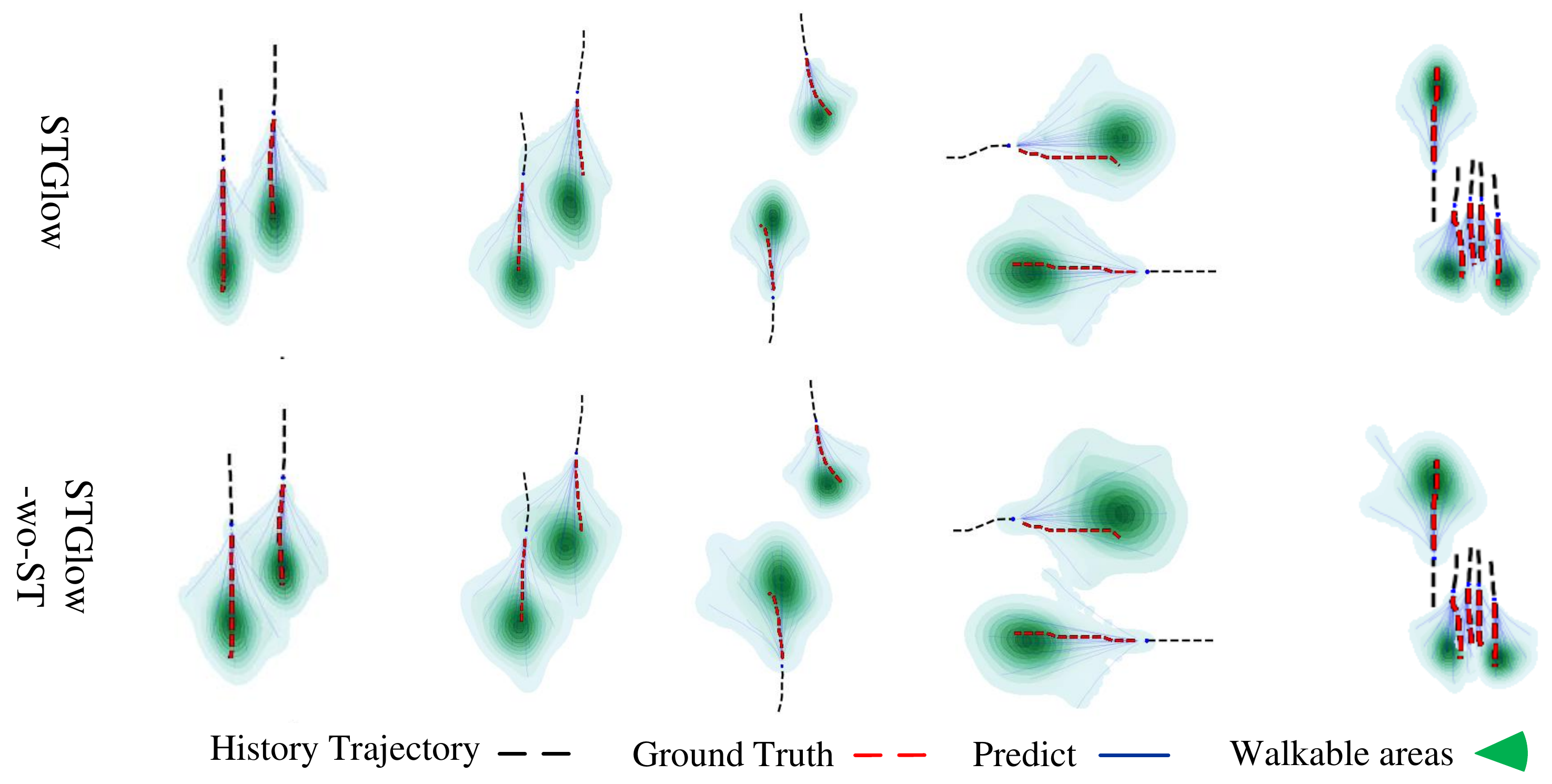}
		\caption{Examples of diverse predictions of our method with and without dual graphormer (ST) on SDD. Black dash represents pedestrian historical trajectory, red dash denotes the ground truth trajectories, thin solid blue line represents the predicted diverse trajectories, and the green area represents the walkable area, which is visualized by the kernel density estimation map drawn from the predicted 20 trajectories. Better viewed in color.}
		\label{fig:diverse}
	\end{figure}
	
	\subsection{Qualitative Analysis}
	In this subsection, we present visual examples to further illustrate the ability of our STGlow to fully explore complex social interactions between pedestrians and generate reasonable and diverse future trajectories.
	
	\subsubsection{Results in different scenarios}
	As can be drawn from Table \ref{tb:comparison}, methods based on CVAEs perform significantly better than methods based on GANs. Thus, we compare the most-likely predictions between STGlow and the previous state-of-the-art CVAE-based method, BiTraP \cite{yao2021bitrap}, qualitatively on all five scenes of the ETH/UCY dataset.
	
	Overall, as shown in Fig. \ref{fig:qual_result}, our prediction results are significantly closer to the ground truth trajectory compared with BiTraP, regardless of simple scenarios or scenarios with complex interactions.  
	Specifically, the first row illustrates simple motion behaviors in five scenarios, including uniform walking and simple interactions. In these scenarios, the trajectories predicted by our method are significantly closer to the ground truth trajectories, since we more precisely model the underlying distribution by optimizing the exact log-likelihood of motion behaviors.
	The second and third rows show more complex motion behaviors, including slowing down, speeding up, making-a-turn, avoiding collision, and complex interactions. In these scenarios, compared to BiTraP, our method is able to make predictions that are more consistent with the laws of human motion, as we intuitively model the evolution of human motion behavior from simple to complex. For example, in the second and third rows of ETH and HOTEL, BiTraP does not handle well with slowing down and speeding up, and even predicts the future trajectory of possible collisions for parallel pedestrians, whereas our method does a good job of forecasting these challenging motion behaviors.
	Besides, our method adequately models temporal dependencies and the mutual spatial interactions to predict a more reasonable future trajectory.
	For instance, in the second row of UNIV, 
	our method can accurately predict the making-a-turn behavior for parallel pedestrians in complex scenarios. In contrast, BiTraP’s prediction results show certain deviations in both velocity and direction.
	Similarly, in the third row, our method can accurately deal with the situation of avoiding collision, because we intuitively model social interactions in both temporal and spatial domain. A similar conclusion can be drawn from the second and third rows of ZARA1 and ZARA2.
	
	\subsubsection{Results of diverse predictions} 
	We further investigate the ability of our method to generate diverse predictions by comparing the case with and without the dual graphormer.
	As shown in the walkable area predicted in Fig. \ref{fig:diverse}, regardless of whether there is social interaction modeling or not, our generative flow-based method can generate diverse and reasonable future trajectories well. Notably, compared with the model without the dual graphormer (as shown in the second row), our STGlow not only avoids collisions to a certain extent but also predicts more concentrated walking areas. This means that our method STGlow considering social interaction can better measure the diversity and stability of forecasted future trajectories.
	\begin{table}[!t]
		\renewcommand{\arraystretch}{1.0}
		\centering  
		\caption{The ADE and FDE performance of variants on SDD dataset. $\bigcirc$ and $\triangle$ indicate GRU and Transformer, respectively }
		\label{components}
		\begin{tabular}{cccc|c}
			\hline
			\multicolumn{4}{c|}{\textbf{Components}}                & \textbf{Variants} \\ \hline
			\textbf{SG} & \textbf{TG} & \textbf{PN} & \textbf{BiD} & \textbf{ADE/FDE}  \\ \hline
			\checkmark           & \checkmark           & \checkmark            &              &7.30/11.38                   \\
			$\times$   &      $\bigcirc$       &              & \checkmark            &8.80/14.75                   \\
			$\times$    &      $\bigcirc$      & \checkmark            & \checkmark            &7.45/11.88                   \\
			$\triangle$    &      $\triangle$      & \checkmark            & \checkmark            &7.33/11.56                   \\
			$\times$    & \checkmark           & \checkmark            & \checkmark            &7.39/11.70                   \\
			\checkmark           &  $\bigcirc$           & \checkmark            & \checkmark            &7.53/11.84                   \\
			\checkmark           & \checkmark           &              & \checkmark            &8.60/14.35                   \\
			\checkmark           & \checkmark           & \checkmark            & \checkmark            &\bf{7.20/11.20}                   \\ \hline
		\end{tabular}
	\end{table}
	
	
	\begin{table}[!t]
		\renewcommand{\arraystretch}{1.0}
		\setlength{\tabcolsep}{1.5mm}
		\centering  
		\caption{The performance of variants of Dual Graphormer on SDD dataset.}
		\label{STG}
		\begin{tabular}{ccc|c|ccc|c}
			\hline
			\multicolumn{3}{c|}{\textbf{TG}}       & \textbf{Results} & \multicolumn{3}{c|}{\textbf{SG}} & \textbf{Results} \\ \hline
			\textbf{CE} & \textbf{PE} & \textbf{$A_{tmp}$} & \textbf{ADE/FDE}     & \textbf{SE}     & \textbf{HE} & \textbf{$A_{spa}$}    & \textbf{ADE/FDE}     \\ \hline
			\textbf{\checkmark}  & $\times$   & $\times$   &7.27/11.42                      & \textbf{\checkmark}      &   $\times$          & $\times$  &7.25/11.35                      \\
			$\times$    & \textbf{\checkmark}  &  $\times$           &7.22/11.43                      &       $\times$           & \textbf{\checkmark}  & $\times$  & 7.25/11.45                     \\
			$\times$     &     $\times$         & \textbf{\checkmark}  &7.24/11.48                   & $\times$       & $\times$   & \textbf{\checkmark}  &7.24/11.36                      \\
			$\times$   &$\times$    & $\times$   &7.30/11.46                      & $\times$       & $\times$   &$\times$  &7.29/11.47                     \\
			\textbf{\checkmark}  & \textbf{\checkmark} & $\times$ & 7.24/11.38  & \textbf{\checkmark}  & \textbf{\checkmark} & $\times$ & 7.23/11.42 \\
			\textbf{\checkmark}  & \textbf{\checkmark}  & \textbf{\checkmark} & \textbf{7.20/11.20} & \textbf{\checkmark}      & \textbf{\checkmark}    & \textbf{\checkmark}  &\textbf{7.20/11.20}  \\ \hline
		\end{tabular}
	\end{table}
	\vspace{-3mm}
	
	\begin{table}[!t]
		\renewcommand{\arraystretch}{1.0}
		\centering  
		\caption{Ablation experiments on hyperparameters of the loss function on SDD dataset.}
		\label{hyperparameter}
		\begin{tabular}{c|l|l|l|l}
			\hline
			$\alpha$ & $\lambda _1$ & $\lambda _2$ & $\lambda _3$ & ADE/FDE \\ \hline
			\multirow{5}{*}{0.0} & 0.0 & 0.0 & 1.0 & 7.25/11.39  \\
			& 0.0 & 1.0 & 0.0 & 7.28/11.50  \\
			& 1.0 & 0.0 & 0.0 & 7.27/11.54  \\
			& 0.25 & 0.25 & 0.5 & 7.23/11.35  \\
			& 0.5 & 0.5 & 0.0 & 7.30/11.47  \\ \hline
			\multirow{5}{*}{0.5} & 0.0 & 0.0 & 1.0 & 7.24/11.30  \\
			& 0.0 & 1.0 & 0.0 & 7.28/11.37  \\
			& 1.0 & 0.0 & 0.0 & 7.26/11.45  \\
			& 0.25 & 0.25 & 0.5 & 7.23/11.28  \\
			& 0.5 & 0.5 & 0.0 & 7.27/11.44  \\ \hline
			\multirow{5}{*}{1.0} & 0.0 & 0.0 & 1.0 & 7.22/11.25  \\
			& 0.0 & 1.0 & 0.0 & 7.27/11.28  \\
			& 1.0 & 0.0 & 0.0 & 7.25/11.40  \\
			& 0.25 & 0.25 & 0.5 & \bf{7.20/11.20}  \\
			& 0.5 & 0.5 & 0.0 & 7.25/11.25  \\ \hline                   
		\end{tabular}
	\end{table}
	\vspace{-3mm}
	
	\begin{table*}[!t]
		\renewcommand{\arraystretch}{1.0}
		\setcounter{table}{6}
		\centering  
		\caption{Comparison of the proposed approaches in terms of inference time.}
		\label{inference_time}
		\begin{tabular}{c|c|c|c|c|c|c}
			\hline
			Methods       & BiTrap         & Trajectorn++ & Y-Net+TTST & AgentFormer & DMRGCN & Ours  \\ \hline
			Time (ms/step) & \textbf{0.014} & 0.864        & 13.294     &  1.882       & 0.852  & \color{red}{\underline{0.174}} \\ \hline
		\end{tabular}
	\end{table*}
	\subsection{Ablation Experiments}
	In this subsection, we conduct ablation experiments to investigate the effectiveness of each key component including spatial graphormer, temporal graphormer, PN in Glow, and bidirectional decoder.
	
	\subsubsection{Componets  of our architecture}
	We first explore the impact of each component of our architecture, including spatial graphormer (SG), temporal graphormer (TG), pattern normalization (PN), and bidirectional decoders (BiD). In the variations of our approach, we choose GRU or Transformer to replace our TG to extract the representation of motion behaviors and use a widely-used forward decoder to replace our BiD. The ablation results are summarized in Table \ref{components}. Obviously, each component in our framework improves performance to some extent. Compared with the third and fifth rows, it can be seen that our TG can extract temporal dependencies better than GRU due to considering the temporal dependencies in both behavior-independent and behavior-dependent situations. Besides, considering the uniqueness of behavior pattern, our PN normalizes the behavior pattern of each pedestrian, bringing a huge performance gain compared to the last two rows. Note that our spatial and temporal graphormer is significantly better than the standard Transformer on modeling social interactions in the fourth row, which further demonstrates the effectiveness of our proposed dual graphormer. Other ablation results also demonstrated the importance of our proposed components.
	
	\subsubsection{Dual graphormer analysis}
	To investigate the impact of each component in the developed temporal graphormer (TG) and spatial graphormer (SG) on trajectory prediction, including the centrality encoding (CE), position embedding (PE), adjacency matrix in temporal ($A_{tmp}$), spatial embedding (SE), steering embedding (HE), and adjacency matrix in spatial ($A_{spa}$), we conducted corresponding ablation experiments which are summarized in Table \ref{STG}.
	As shown by the left and right groups in the table, in general, each component we designed in TG and SG help to better model social interactions in both temporal and spatial domains, resulting in improved performance. Note that when no components are adopted, our TG and SG degenerate into commonly used Transformers, where the time step nodes and pedestrians in the scene are treated as fully connected undirected graphs. Obviously, 
	our SG and TG significantly outperform the widely used Transformer that does not consider temporal dependencies and mutual spatial interactions.
	
	\subsubsection{Hyperparameters analysis}
	To ensure the rationality of the hyperparameter settings in the loss function $L_{traj}$, we further performed ablation experiments for hyperparameters $\alpha$, $\lambda _1$, $\lambda _2$ and $\lambda _3$. The experimental results are summarized in Table \ref{hyperparameter}. Since the forward and backward trajectory predictions have the same importance, we generally keep the balance coefficients (i.e., $\lambda _1$ and $\lambda _2$) consistent for both in ablation experiments. Note that when $\lambda _1 = 1.0$, we adopt the forward predicted trajectories as the final trajectories. Similarly, when $\lambda _2 = 1.0$, the backward predicted trajectories served as the final trajectories. In addition, we adopt the forward predicted trajectories as the final trajectories when $\lambda _1 = \lambda _2 = 0.5, \lambda _3 = 0.0$ and the bidirectional predicted trajectories as the final trajectories when $\lambda _3 \neq 0.0$.
	As shown in Table \ref{hyperparameter}, in general, each part of the loss function $L_{traj}$ (\textit{i.e.}, goal estimation, forward trajectory prediction, backward trajectory prediction, and bidirectional trajectory prediction) contributes to the proposed model. Furthermore, we observed that bidirectional trajectory prediction ($\lambda _3 = 1.0$) is slightly superior to forward trajectory prediction ($\lambda _1 = 1.0$) and backward trajectory prediction ($\lambda _2 = 1.0$). When supervising unidirectional and bidirectional trajectory prediction simultaneously, the model achieves the best performance.
	Based on the experimental results, we finally adopt: $\alpha = 1.0$, $\lambda _1 = 0.25$, $\lambda _2 = 0.25$ and $\lambda _3 = 0.5$ in our experiments.
	\subsection {Inference Time Analysis}
	To verify the efficiency of our proposed method, we conduct a comparison experiment on inference time with existing mainstream trajectory prediction frameworks. As demonstrated in Table \ref{inference_time}, our method is inferior only to BiTrap in inference time but has significant performance improvements.
	\begin{figure}[!t]
		\centering
		\includegraphics[width=0.95\linewidth]{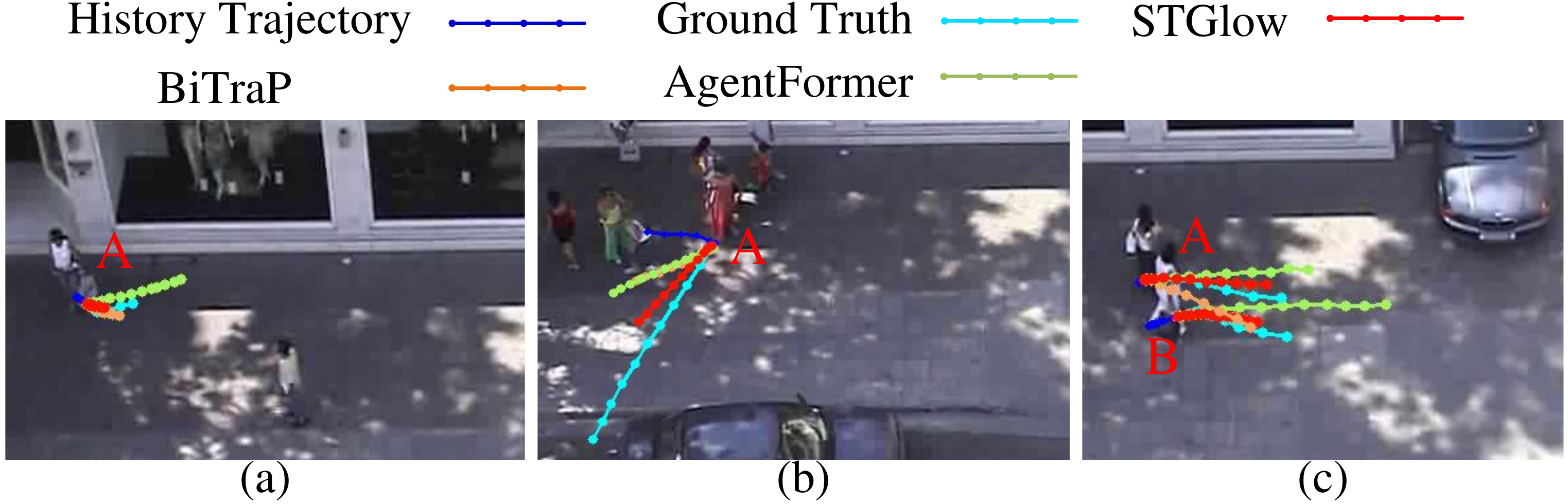}
		\caption{Cases with poor performance on the ETH/UCY dataset. Better viewed in color.}
		\label{fig:discussion}
	\end{figure}
	
	\subsection{Discussion}
	In this subsection, we discuss some limitations of our approach. First, we experimentally find that our method fails to accurately predict the scenes when the motion behaviors of pedestrians change drastically in a short period of time. 
	Fig. \ref{fig:discussion} provides several examples where pedestrians abruptly transition their motion behaviors. Specifically, pedestrian A in Fig. (a) was in a standing state for the first 6 time steps within the 8 observed time steps, and started to move slowly in the last two time steps. In Fig. (b), pedestrian A walked to the right at a constant speed for the first 5 time steps of the observation, but suddenly reduced her speed and changed her walking direction in the last three time steps. In Fig. (c), pedestrian B walked at a very slow speed for the first 7 time steps of the observation and then started walking at a normal speed in the last time step. Meanwhile, pedestrian A was stationary until the last time step of the observation, when it suddenly began walking.
	Though our approach can still cope with such scenarios somewhat better than existing methods such as BiTrap and AgentFormer, it still has a large error compared to the ground truth trajectory. How to address the scenes when the motion behaviors of pedestrians change drastically still needs more effort in future research.
	
	Besides, modeling complex social interactions between pedestrians increases the time cost in the inference phase. When evaluated on ETH/UCY datasets, our method required 0.174 ms/step, BiTrap required 0.014 ms/step, Trajectron++ required 0.864 ms/step, and DMRGCN required 0.852 ms/step. Although our method is faster than Trajectron++ and DMRGCN, it is much slower than BiTrap. Fortunately, there has been a lot of recent work focusing on improving the efficiency of Transformers \cite{3530811, choromanski2021rethinking}. We leave it as future work to build more efficient interaction modules.
	\section{Conclusion}  \label{sec:conclusion}
	In this paper, we have introduced a novel STGlow framework for trajectory prediction. Different from previous approaches, our method can more precisely model the underlying data distribution by optimizing the exact log-likelihood of observations. Besides, our method has clear physical meanings to simulate the evolution of human motion behaviors, where the forward process of the flow gradually decouples the complex motion behavior into a series of simple behaviors, while its reverse process represents the evolution of simple behaviors to the complex motion behavior.
	In addition, we have designed a novel dual graphormer to extract the global social interaction of pedestrians in both temporal and spatial domains. Both quantitative and qualitative experimental results demonstrate the superiority of our approach under various situations.

	\ifCLASSOPTIONcaptionsoff
	\newpage
	\fi
	
	\bibliography{STGlow}
	\bibliographystyle{IEEEtran}
	\vspace{-15 mm}
	\begin{IEEEbiography}
		[{\includegraphics[width=1in,height=1.25in,clip,keepaspectratio]{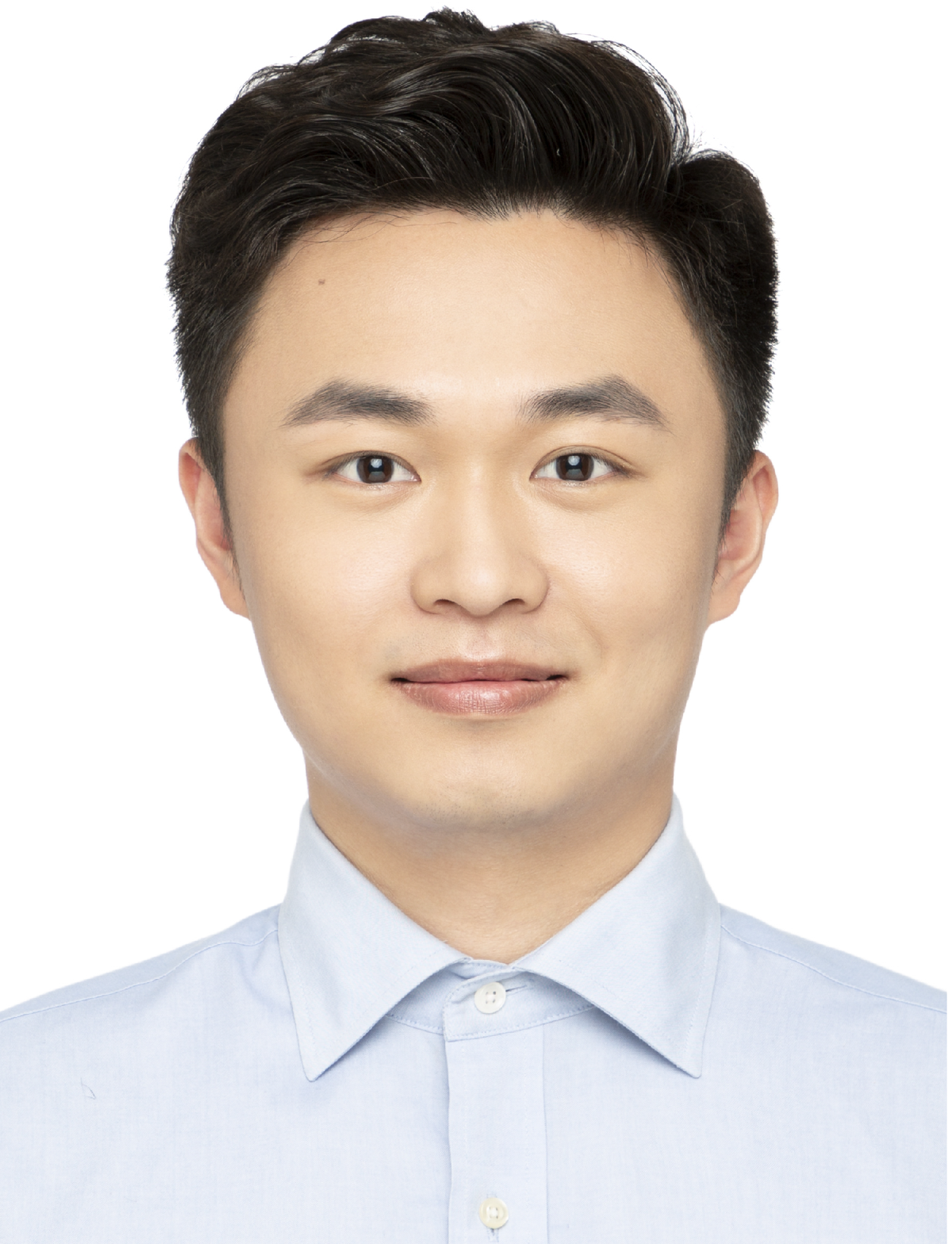}}]{Rongqin Liang} (Student Member, IEEE) received the B.Eng. degree in communication engineering from Wuyi University, Guangdong, China, in 2018 and M.S. degree in Information and Communication Engineering from Shenzhen University, Shenzhen, China, in 2021.
		He is currently a Ph.D. candidate at the College of Electronics and Information Engineering from Shenzhen University. His current research interests include trajectory prediction, anomaly detection, computer vision and deep learning.
	\end{IEEEbiography}
	\vspace{-15 mm} 
	\begin{IEEEbiography}
		[{\includegraphics[width=1in,height=1.25in,clip,keepaspectratio]{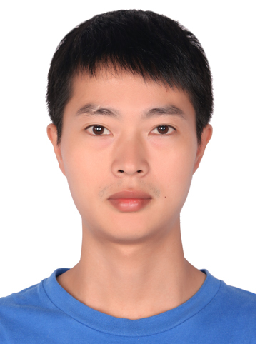}}]{Yuanman Li} (Member, IEEE) received the B.Eng. degree in software engineering from Chongqing University, Chongqing, China, in 2012, and the Ph.D. degree in computer science from University of Macau, Macau, 2018. From 2018 to 2019, he was a Post-doctoral Fellow with the State Key Laboratory of Internet of Things for Smart City, University of Macau. He is currently an Assistant Professor with the College of Electronics and Information Engineering, Shenzhen University, Shenzhen, China. His current research interests include multimedia security and forensics, data representation, computer vision and machine learning.
	\end{IEEEbiography}
	\vspace{-15 mm} 
	\begin{IEEEbiography}
		[{\includegraphics[width=1in,height=1.25in,clip,keepaspectratio]{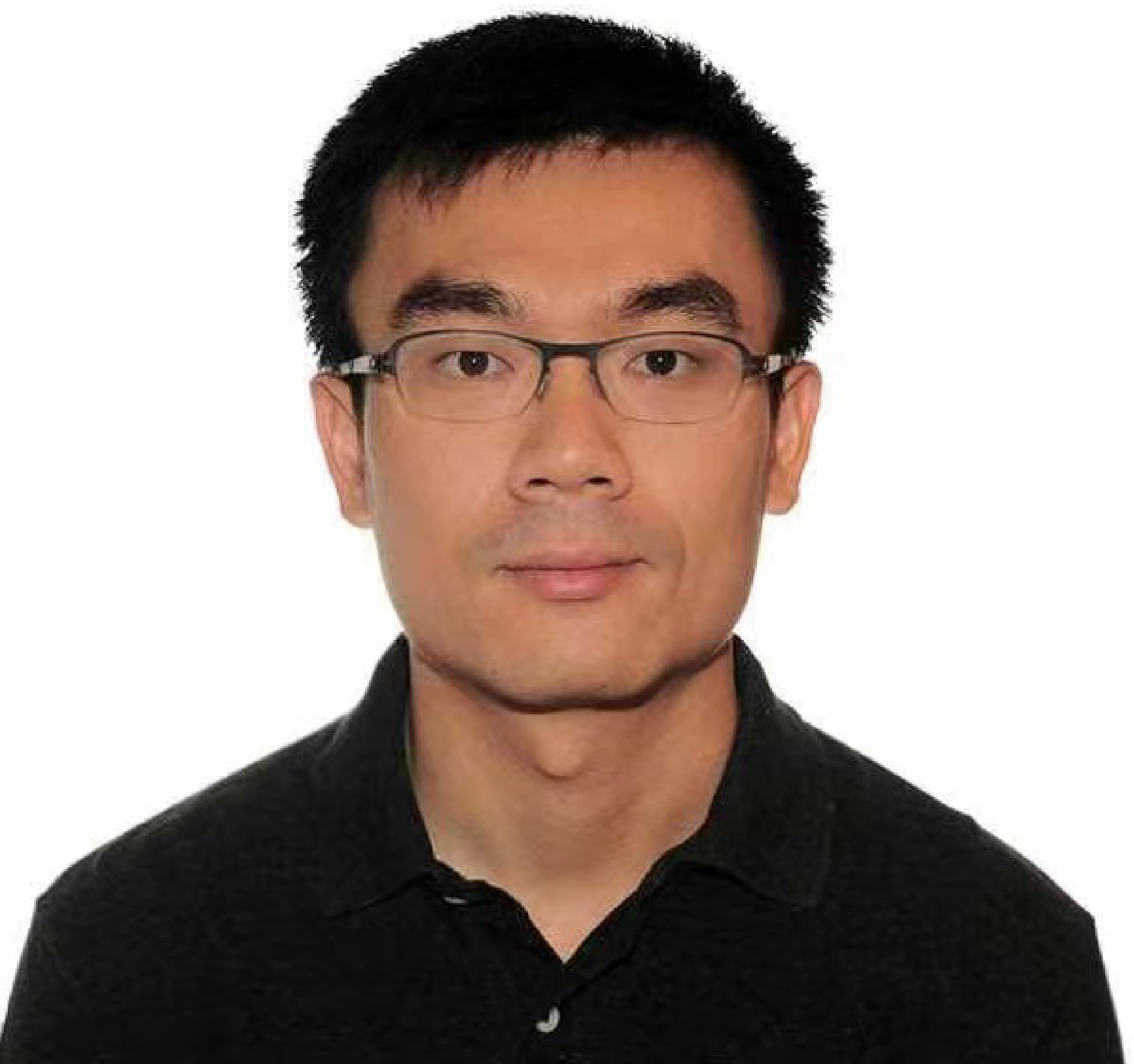}}]{Jiantao Zhou} (Senior Member, IEEE) received the B.Eng. degree from the Department of Electronic Engineering, Dalian University of Technology, in 2002, the M.Phil. degree from the Department of Radio Engineering, Southeast University, in 2005, and the Ph.D. degree from the Department of Electronic and Computer Engineering, Hong Kong University of Science and Technology, in 2009. He held various research positions with University of Illinois at Urbana-Champaign, Hong Kong University of Science and Technology, and McMaster University. He is an Associate Professor with the Department of Computer and Information Science, Faculty of Science and Technology, University of Macau, and also the Interim Head of the newly established Centre for Artificial Intelligence and Robotics. His research interests include multimedia security and forensics, multimedia signal processing, artificial intelligence and big data. He holds four granted U.S. patents and two granted Chinese patents. He has co-authored two papers that received the Best Paper Award at the IEEE Pacific-Rim Conference on Multimedia in 2007 and the Best Student Paper Award at the IEEE International Conference on Multimedia and Expo in 2016. He is serving as the Associate Editors of the IEEE TRANSACTIONS on IMAGE PROCESSING and the IEEE TRANSACTIONS on MULTIMEDIA.
	\end{IEEEbiography}	
	\vspace{-15 mm} 
	\begin{IEEEbiography}
		[{\includegraphics[width=1in,height=1.25in,clip,keepaspectratio]{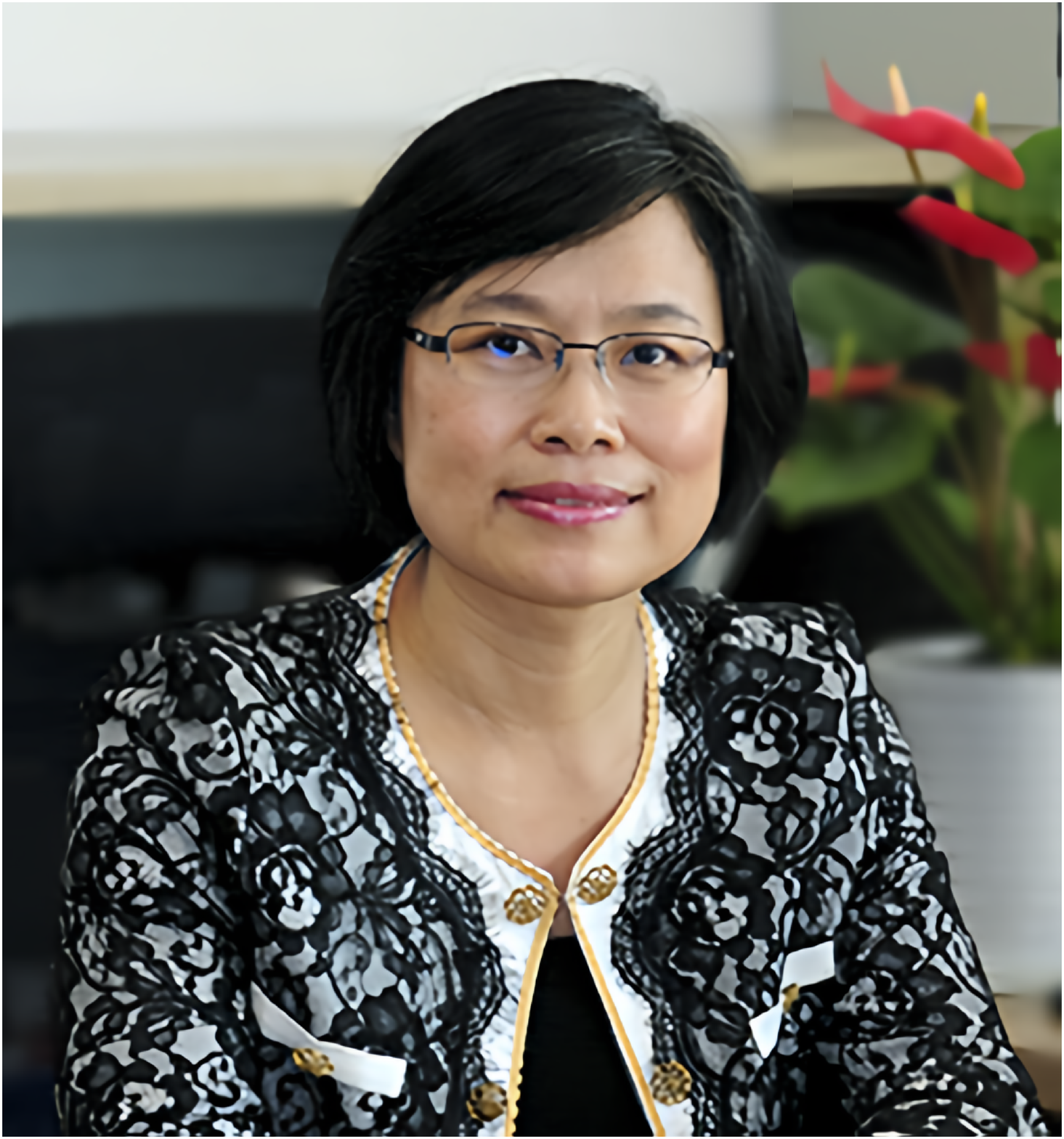}}]{Xia Li} (Member, IEEE) received her B.S. and M.S. in electronic engineering and SIP (signal and information processing) from Xidian University in 1989 and 1992 respectively. She was later conferred a Ph.D. in Department of information engineering by the Chinese University of Hong Kong in 1997. Currently, she is a member of the Guangdong Key Laboratory of Intelligent Information Processing. Her research interests include intelligent computing and its applications, image processing and pattern recognition.
	\end{IEEEbiography}
	
\end{document}